\documentclass[11pt,a4paper]{article}
\usepackage{times,latexsym}
\usepackage{url}
\usepackage[T1]{fontenc}
\usepackage[most]{tcolorbox}
\definecolor{casegreen}{RGB}{58,118,76}
\definecolor{casered}{RGB}{170,60,60}
\definecolor{caseblue}{RGB}{55,90,150}
\definecolor{casegray}{RGB}{245,246,248}
\definecolor{caseborder}{RGB}{80,80,80}

\usepackage{tikz}

\newcommand{\casebadge}[1]{%
\tikz[baseline=(X.base)]{
\node[
    fill=black,
    text=white,
    rounded corners=5pt,
    inner xsep=12pt,
    inner ysep=5pt,
    font=\bfseries\large
] (X) {#1};
}%
}

\usepackage[acceptedWithA]{tacl2021v1}

\usepackage{xspace,mfirstuc,tabulary}

\newif\iftaclinstructions
\taclinstructionsfalse 
\iftaclinstructions
\renewcommand{\confidential}{}
\renewcommand{\anonsubtext}{(No author info supplied here, for consistency with
TACL-submission anonymization requirements)}
\newcommand{\instr}
\fi

\iftaclpubformat 

\else

\fi
\usepackage{times}
\usepackage{latexsym}
\usepackage{longtable}
\usepackage[T1]{fontenc}

\usepackage[utf8]{inputenc}
\usepackage{graphicx}
\usepackage{microtype}

\usepackage{times}  
\usepackage{graphicx} 
\usepackage{natbib}  
\usepackage{algorithm}
\usepackage{array}
\usepackage{latexsym}

\usepackage[T1]{fontenc}

\usepackage[utf8]{inputenc}
\usepackage{multirow}
\usepackage{microtype}
\usepackage{url}
\usepackage{amsmath}
\usepackage{amssymb}
\usepackage{color,xcolor,colortbl}
\usepackage{enumitem}
\usepackage{longtable}

\usepackage{algorithm}
\usepackage{algpseudocode}
\usepackage{amsmath}
\usepackage{wrapfig}
\usepackage{bm}
\usepackage{booktabs}
\usepackage{mathtools}
\usepackage{array}
\usepackage{pbox}
\usepackage{pifont}
\usepackage{tabularx}
\usepackage{tablefootnote}
\usepackage{float}
\usepackage{booktabs}
\usepackage{arydshln}

\usepackage{amsmath,amsfonts,bm}

\def\eqref#1{equation~\ref{#1}}

\def\1{\bm{1}}

\DeclareMathAlphabet{\mathsfit}{\encodingdefault}{\sfdefault}{m}{sl}
\SetMathAlphabet{\mathsfit}{bold}{\encodingdefault}{\sfdefault}{bx}{n}

\usepackage{hyperref}
\usepackage{url}
\hypersetup{
  pdftitle={MultiViewDx: Evidence-Linked Multi-View Clinical Diagnosis},
  pdfauthor={Junda Wang, Zonghai Yao, Yujan Ting, Eric Z. Chen, Hieu Tran, Hong Yu, Weijing Huang, and Terrence Chen}
}

\newcommand{\revmod}[1]{#1}
\newcommand{\revadd}[1]{#1}
\newcommand{\revaddcolor}{}

\newenvironment{revaddblock}{\begingroup}{\endgroup}
\title{MultiViewDx: Evidence-Linked Multi-View Clinical Diagnosis}

\author{
\textbf{Junda Wang}$^{1,2}$,
\quad \textbf{Zonghai Yao}$^{2,3}$,
\quad \textbf{Yujan Ting}$^{1}$,
\quad \textbf{Eric Z. Chen}$^{1}$, \\
\textbf{Hieu Tran}$^{1,2}$,
\quad \textbf{Hong Yu}$^{2,3}$,
\quad \textbf{Weijing Huang}$^{1}$,
\quad \textbf{Terrence Chen}$^{1}$ \\
$^{1}$United Imaging Intelligence, MA, USA \\
$^{2}$Manning College of Information and Computer Sciences, UMass Amherst, MA, USA \\
$^{3}$Miner School of Computer and Information Sciences, Umass Lowell, MA, USA
}

\begin{document}

\maketitle

\begin{abstract}

Medical multimodal large language models (MLLMs) can perform well on existing medical visual question answering (MedVQA) benchmarks, but their training data often does not match clinical diagnosis. Most supervision is organized around isolated images or short QA pairs, leaving two structures weakly specified: how evidence leads to a decision, and how views, series, modalities, and patient context from the same case are linked. We introduce MultiViewDx, a partly physician-validated multimodal instruction dataset for evidence-linked multi-view medical imaging diagnosis. MultiViewDx uses the clinical case as the supervision unit. It links imaging studies with patient context, normalizes heterogeneous reports into an evidence-linked workflow (evidence → findings → differential discussion → diagnosis), and uses a unified image-text retriever to constrain instruction synthesis to source-supported evidence. It covers X-ray, CT, MRI, ultrasound, histopathology, and other clinical visual sources. We fine-tune MultiViewDx-8B-AN and evaluate it on both existing MedVQA benchmarks and real-world case-based diagnostic reasoning. Across four MedVQA benchmarks, it achieves the best average accuracy among compared systems (79.0\%), outperforming HuatuoGPT-Vision-34B (66.7\%) and Claude3-Opus (55.7\%). Beyond MedVQA, on JAMA Clinical Challenge cases, it receives the strongest overall rating under a physician-designed rubric for key clinical points, diagnostic inference, and evidence grounding. Controlled ablations and clinician evaluation show that both case-level multi-view organization and evidence-linked reasoning targets contribute to the gain.
\footnote{Code and datasets are available at \url{https://github.com/believewhat/MultiViewDx}.}

\end{abstract}

\section{Introduction}

Medical imaging diagnosis in clinical practice is rarely a single-image, single-answer task. A clinician usually reviews a study as a set of related views, slices, or sequences, and interprets them with patient history, laboratory results, vital signs, medications, and other clinical signals. The one or two images shown in a case report, teaching example, or patient-facing explanation are selected for communication; they are not necessarily the full evidence used in care. When such material is converted into one-image QA examples, the task can become visual recognition or answer matching rather than case-level diagnosis.

Medical multimodal large language models (MLLMs) have improved medical image QA through biomedical image-text corpora and instruction tuning~\citep{yan2023multimodal, liu2024visual, jin2024hidden, yang2025unveiling, li2024llava, chen2024huatuogpt, yao2025survey}. Recent systems such as LLaVA-Med, HuatuoGPT-Vision, MedTrinity, and Hulu-Med adapt general MLLMs to medical data~\citep{li2024llava, chen2024huatuogpt, xie2024medtrinity, jiang2025hulu}. However, much of the supervision still uses isolated image-caption pairs or short QA instances. We study \emph{evidence-linked multi-view diagnosis}, where the clinical case is the training and evaluation unit. The model must identify findings across available views, slices, sequences, or modalities, connect them to patient context, compare differentials, and state a diagnosis or next step with explicit evidence support.

\begin{figure*}[!ht]
    \centering
    \includegraphics[width=0.9\textwidth]{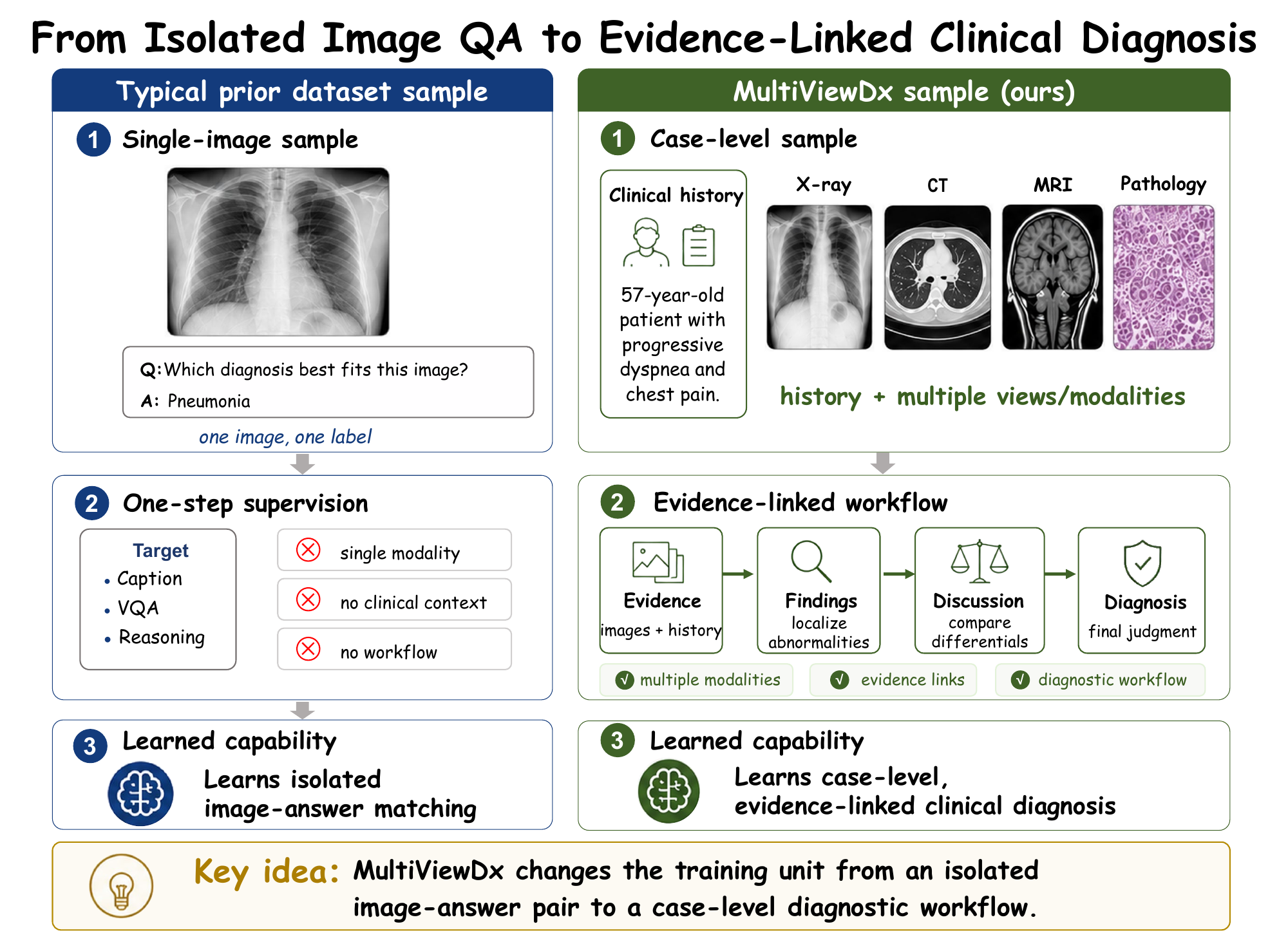}
    \caption{\textbf{From image-level QA to case-level diagnosis.} Left: common medical VQA or caption data use one displayed image and a short target answer. Right: MultiViewDx uses the clinical case as the training unit, linking patient context, multiple views or series, localized evidence, intermediate findings, differential discussion, and the final diagnosis or next step. The key change is the supervision unit: from an isolated image-answer pair to an evidence-linked diagnostic case.}
\label{fig:overview}
\end{figure*}

Current datasets often leave this structure under-specified in two ways. \textbf{Workflow-sparse supervision} omits the diagnostic path from evidence to findings, differential discussion, and final decision; image captions and short QA pairs usually state what is present, but not how it supports a clinical conclusion~\citep{bustos2020padchest, irvin2019chexpert, johnson2019mimic, ikezogwo2024quilt}. \textbf{Case-fragmented supervision} treats views or series from the same study as independent samples, or drops them entirely. This simplification is understandable, because public reports and case presentations often keep only a few illustrative images. But it makes the data easier than clinical reasoning, since the original diagnostic process used a richer case-level evidence set~\citep{xie2024medtrinity, pal2023med, dorfner2024biomedical}.

This mismatch is hard to see if evaluation stops at existing MedVQA benchmarks. Benchmarks such as VQA-RAD, SLAKE, PathVQA, and PMC-VQA remain useful for measuring \revmod{medical QA performance}, but many items can be answered from a single image or a short visual-text association. They do not consistently require cross-view corroboration, evidence attribution, or integration of patient context. We therefore report both standard MedVQA performance and diagnosis-oriented evaluation on JAMA Clinical Challenge cases, using rubric-based scoring and clinician review.

To address these limitations, we introduce \textbf{MultiViewDx}, a data-centric instruction corpus constructed from real, de-identified clinical data with physician validation over sampled synthesized reasoning chains. MultiViewDx changes the supervision unit from an isolated image-answer pair to an evidence-linked clinical case. It links imaging studies to clinical context, preserves available view and series structure, and normalizes heterogeneous reports into a workflow of evidence, findings, differential discussion, and diagnosis or next step. A unified image-text retriever gathers source-supported image and context evidence before instruction synthesis. Each instance is designed to teach both the answer and the provenance of the key evidence.

We fine-tune MultiViewDx-8B-AN and test whether this data design changes model behavior. On VQA-RAD and SLAKE, MultiViewDx-8B-AN achieves the best average accuracy among compared models, \revmod{showing strong performance on existing MedVQA benchmarks}. On JAMA case-based diagnostic reasoning, it obtains the strongest overall rating under a physician-designed rubric for key clinical points, diagnostic inference, and evidence grounding. The ablations tie these gains to the proposed design: removing case-level multi-view organization substantially reduces clinician-rated quality, while removing explicit reasoning targets further weakens clinician judgments. Physician validation of synthesized reasoning chains and clinician evaluation of model outputs provide checks beyond automatic answer matching.

\paragraph{Contributions.}
\begin{itemize}
    \item MultiViewDx, an evidence-linked multi-view instruction dataset that uses the clinical case as the supervision unit and aligns imaging views, patient context, and staged diagnostic fields
    \item a retrieval-constrained construction pipeline that preserves evidence provenance across views, regions, and clinical signals during instruction synthesis
    \item an evaluation and ablation protocol that combines existing MedVQA benchmarks with real-world case-based diagnostic reasoning, rubric scoring, and clinician evaluation to \revmod{distinguish benchmark QA performance from diagnostic reasoning}.
\end{itemize}

\section{Related Work}

\paragraph{Existing Multimodal Medical Datasets}
Medical multimodal datasets have evolved from report-paired corpora such as MIMIC-CXR-JPG~\citep{johnson2019mimic} to large image-caption collections like PMC-OA~\citep{lin2023pmc}, case/VQA resources (PMC-CaseReport, PMC-VQA)~\citep{wu2023towards, zhang2023pmc}, instruction-style sets from PubMed imagery (LLaVA-Med VQA, PubMedVision)~\citep{li2024llava, chen2024huatuogpt}, structured annotation efforts (RadGenome-Chest CT)~\citep{zhang2024radgenome}, and multigranular pipelines such as MedTrinity~\citep{xie2024medtrinity}, Hulu-Med Data~\citep{jiang2025hulu}, Lingshu Data~\citep{xu2025lingshu} and HealthGPT Data~\citep{lin2025healthgpt}. 
Despite this progress, much of the available supervision remains image-level: a displayed image is paired with a caption, a short answer, or a generated instruction.
This creates workflow-sparse supervision because intermediate diagnostic fields and evidence provenance are not consistently represented.
It also creates case-fragmented supervision because views, slices, sequences, and non-imaging signals from the same clinical study are often separated into independent examples.
Recent resources that include multi-view inputs or richer clinical context, such as MedThinkVQA~\citep{yaomedical} and MedXpertQA~\citep{zuo2025medxpertqa}, are valuable for evaluation but remain too small to train large medical MLLMs at scale.
MultiViewDx differs by organizing training data around evidence-linked clinical cases: it aligns multi-view studies with patient context, normalizes reports into staged diagnostic fields, and constrains instruction synthesis with retrieved source evidence.

\paragraph{Medical Multimodal Models}
Several efforts fine-tune general-purpose multimodal models on medical data and show strong gains.
Med-Flamingo~\citep{moor2023med} adapts OpenFlamingo-9B~\citep{chen2024visual} with a medical corpus, and Med-PaLM~\citep{tu2024towards} extends PaLM-E~\citep{driess2023palm} using large-scale medical image-text pairs. 
Similarly, LLaVA-Med, Med-Gemini~\citep{saab2024capabilities}, HuatuoGPT-Vision~\citep{chen2024huatuogpt}, HealthGPT~\citep{lin2025healthgpt}, Lingshu~\citep{xu2025lingshu} and Hulu-Med~\citep{jiang2025hulu} leverage instruction tuning over curated PubMed-derived image-text data for medical QA.
Our focus is complementary to model-architecture work.
We do not propose a new fusion module; instead, we study how changing the supervision unit from isolated images to evidence-linked clinical cases affects both \revmod{held-out MedVQA performance} and case-based diagnostic reasoning.

\section{MultiViewDx}

MultiViewDx is built around a simple design choice: the clinical case, rather than the isolated image, is the basic training unit.
For each case, we preserve two relations when they are available.
The first is \emph{vertical provenance}: evidence is linked to localized findings, differential discussion, and the final diagnosis or next step.
The second is \emph{horizontal study integrity}: related views, slices, sequences, modalities, and non-imaging clinical signals remain grouped under the same case.
This design directly targets workflow-sparse supervision and case-fragmented supervision.

\subsection{Data Collection}

\begin{figure}[t]
    \centering
    \includegraphics[width=\linewidth]{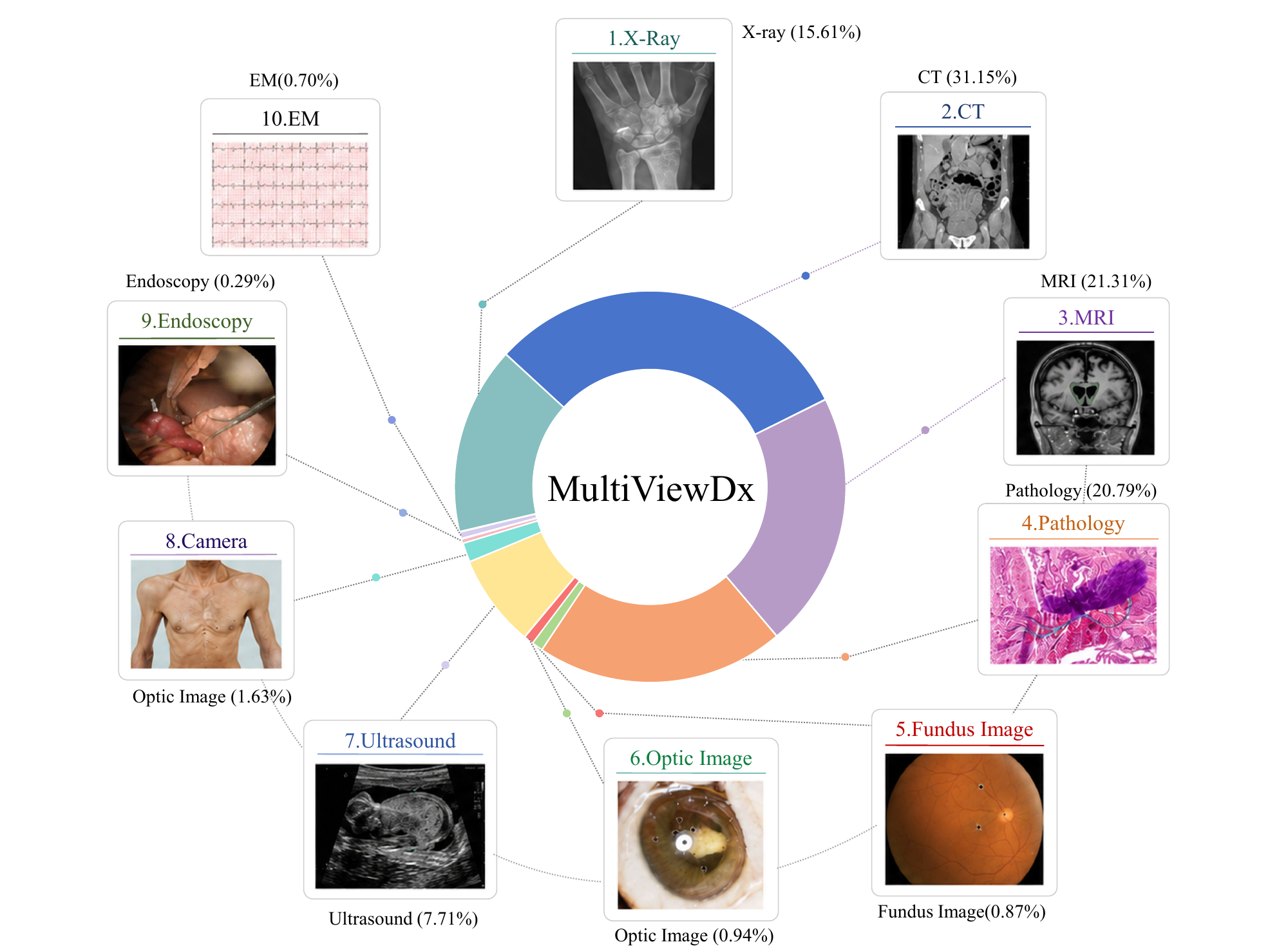}
    \caption{\textbf{Modality coverage of MultiViewDx.} MultiViewDx covers a broad range of medical imaging modalities and clinical visual sources, including X-ray, CT, MRI, histopathology, ultrasound, endoscopy, clinical photography, fundus/optic imaging, and electrophysiology. The central distribution summarizes the major modality composition of the dataset, with CT, MRI, histopathology, and X-ray forming the largest proportions. This diverse modality coverage supports training on multi-view and multi-modal clinical evidence rather than a single image type. }
\label{fig:modality}
\end{figure}

\paragraph{Data sources and study preservation.}
We collect data from sources that provide complementary forms of medical supervision, including report-paired radiology data, pathology VQA, public biomedical image-text corpora, and expert-authored clinical case repositories.
When a source contains study structure, such as AP/LAT radiographs, CT slices, MRI sequences, or ROI hints, we preserve this structure under a shared case identifier instead of converting each image into an unrelated example.
Fig.~\ref{fig:modality} summarizes the resulting modality coverage, with CT, MRI, histopathology, and X-ray forming the largest portions, and additional coverage from ultrasound, endoscopy, clinical photography, fundus or optic imaging, and electrophysiology.

\paragraph{Image and slice selection.}
For 3D CT and MRI studies, we represent each study with a compact set of informative 2D slices.
We retain clinically marked slices when available and uniformly sample additional slices along the study axis, with at most 20 slices per case.
This cap is intended to preserve anatomical coverage and cross-slice continuity while reducing redundancy from densely adjacent slices under limited visual-context budgets.
The all-slice comparison in Sec.~\ref{sec:ablation} tests whether simply exposing more slices is sufficient.

\paragraph{Data Source}
\label{data_source_section}
\leavevmode\revmod{The training corpus draws from} DeepLesion, MIMIC-CXR-JPG, PadChest, Quilt, LLD-MMRI, and MAMA-MIA, along with benchmark training QA datasets such as VQA-RAD, PathVQA, PMC VQA, and SLAKE, covering multiple modalities like CT, MRI, X-ray and so on. Additionally, we expanded the dataset with data from Eurorad and Radiopaedia to include more diverse modalities as shown in table~\ref{tab:data_source_merged}. To support bilingual instruction following, we uniformly sample 300k instances from the combined training corpus and translate their textual fields into Chinese. The translated examples preserve the source image, modality, annotation metadata, and provenance information when available.

\begin{table}[!ht]
\centering
\scriptsize
\setlength{\tabcolsep}{2.3pt}
\renewcommand{\arraystretch}{0.92}
\resizebox{\linewidth}{!}{%
\begin{tabular}{l r c c c c c l}
\toprule
\textbf{Dataset} & \textbf{N} & \textbf{Mod} & \textbf{ROI} & \textbf{HA} & \textbf{Slice} & \textbf{Cap} & \textbf{Lic (short)} \\
\midrule
DeepLesion      & 24,821    & CT    & $\times$      & $\times$         & $\times$ & Y & CC-BY4 \\
PadChest        & 150,730   & XR    & $\times$      & $\checkmark$     & --       & Y & AGR-PadChest \\
Eurorad         & 691,370   & Multi & $\checkmark$  & $\checkmark$     & $\checkmark$ & Y & CC-BY4 \\
MIMIC-CXR-JPG   & 620,113   & XR    & $\times$      & $\checkmark$     & --       & N & PN-CHDL1.5 \\
LLD             & 30,390    & MR    & $\checkmark$  & $\times$         & $\checkmark$ & Y & AGR-LLD \\
MAMA-MIA        & 76,381    & MR    & $\checkmark$  & $\times$         & $\checkmark$ & Y & CC-BY-NC-SA4 \\
PMC-VQA$^{*}$   & 152,603   & Multi & $\times$      & $\checkmark$     & --       & Y & CC-BY-SA \\
Path-VQA        & 19,654    & Path  & $\times$      & $\checkmark$     & --       & Y & CC-BY4 \\
PMC-Instruct    & 619,606   & Multi & $\times$      & $\checkmark$     & --       & Y & OpenRAIL \\
Quilt           & 1,017,416 & Histo & $\times$      & $\checkmark$     & --       & Y & -- \\
Radiopaedia     & 1,131,614 & Multi & $\checkmark$  & $\checkmark$     & $\checkmark$ & N & AGR-RP \\
SLAKE           & 9,835     & Multi & $\times$      & $\checkmark$     & --       & Y & CC-BY4 \\
VQA-RAD         & 1,798     & Multi & $\times$      & $\checkmark$     & --       & Y & CC-BY4 \\
AMBOSS \& JAMA  & 45,820    & Mix   & $\checkmark$  & $\checkmark$     & --       & N & AGR-JAMA \\
\bottomrule
\end{tabular}%
}
\caption{
Unified data sources with annotation, caption availability, and license.
\revmod{N denotes processed sample records. For fragmented sources, the retained metadata support sample-level counts only; unique patient-, study-, and case-level counts are unavailable.}
Mod: CT=CT, XR=X-ray, MR=MRI, Path=pathology, Histo=histopathology, Multi=multi-modality, Mix=multi + text.
ROI: ROI annotation.
HA: human annotation.
Slice: per-slice identifier.
Cap: caption availability (Y/N).
Lic (short): license type
(CC-BY4, CC-BY-SA, CC-BY-NC4, CC-BY-NC-SA4, PN-CHDL1.5, OpenRAIL, AGR-* = dataset-specific agreement).
“--” means not applicable or not provided.
\revadd{$^{*}$PMC-VQA and the Stage~1 corpus share PMC/PMC-OA source provenance and may contain overlapping articles, images, or captions.}
}
\label{tab:data_source_merged}
\end{table}

\noindent \textbf{Human Annotated Workflow} To supervise the end-to-end diagnostic workflow \textbf{(Workflow-sparse supervision)}, we prioritize expert–labeled sources (e.g., Eurorad, Radiopaedia) that mirror real workflows (history $\rightarrow$ findings $\rightarrow$ differential $\rightarrow$ impression/plan). Because raw annotations vary in length and style, we standardize them into consistent workflow fields and align sentences to views/ROIs. For lengthy reports (e.g., Eurorad), we decompose them into (i) per-image findings, (ii) study-level synthesis, and (iii) discussion/decision text, regenerating only format (not content) to preserve expert intent. \revadd{Only explicit figure, panel, view, slice, or ROI references create image-level links; all other statements remain at study level. Specific-image text maps to per-image findings, cross-image integration maps to study-level synthesis, and diagnosis, differential, impression, management, or next-step text maps to discussion/decision content.} The result is study-level supervision that simultaneously (i) teaches stepwise reasoning and decisions (addresses Workflow-sparse supervision) and (ii) ties claims to specific views/slices and clinical context (supports Case-fragmented supervision). A concrete human-annotated case example is provided in Appendix~\ref{apx:human_case_example}.

\begin{figure*}[t]
    \centering
    \includegraphics[width=1.0\textwidth]{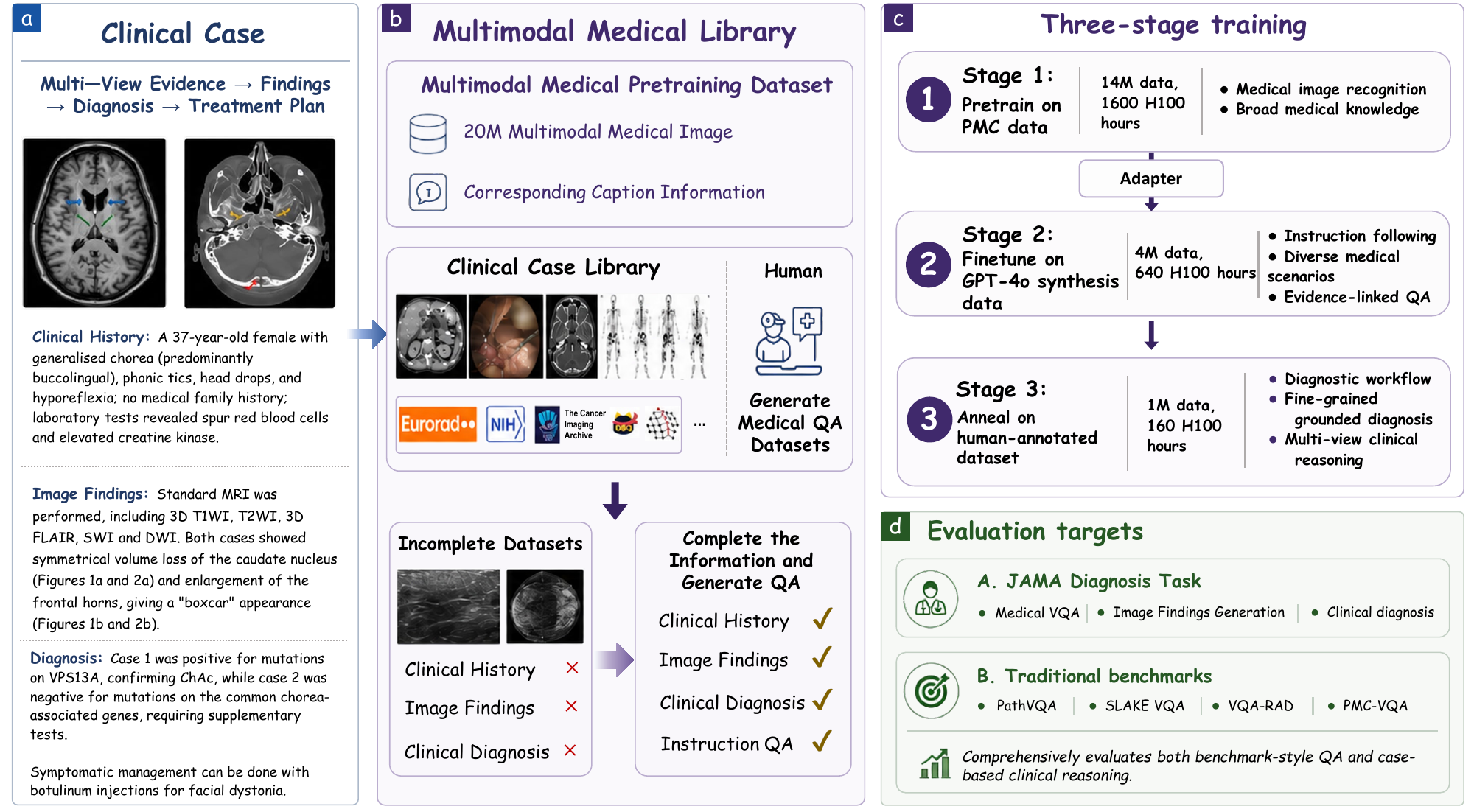}
    \caption{ \textbf{MultiViewDx data construction and training pipeline.} The pipeline starts from clinical cases that contain multi-view evidence, image findings, diagnosis, and treatment-related information. These cases are organized into a multimodal medical library that combines large-scale image-caption pretraining data with a clinical case library. For incomplete sources, retrieval and GPT-4o synthesis are used to complete missing clinical history, image findings, diagnosis, and instruction QA. The model is then trained in three stages: pretraining on PMC image-text data, fine-tuning on GPT-4o-synthesized instruction data, and sequential fine-tuning on human-annotated clinical cases. The resulting model is evaluated on both real-world medical tasks and traditional medical VQA benchmarks. }

\label{fig:pipeline}
\end{figure*}

\subsection{Data Construction Pipeline}

\paragraph{Case preservation and normalization.}
For each source case, we first assign a case identifier and collect all available images, view or slice metadata, report text, ROI hints, and clinical signals.
We normalize modality names, laterality, anatomy terms, and report sections into a shared schema.
Near-duplicate cases are removed by matching source identifiers and by checking high-overlap image-text pairs.
The normalized fields are clinical context, visual evidence, localized image findings, differential discussion, diagnosis or impression, next step, and evidence citations.

\paragraph{Retrieval-constrained synthesis.}
For incomplete cases, we use a unified image-text retriever to retrieve source-supported context before synthesis.
The retriever indexes clinical guidelines and image-report cases, and returns a small set of relevant items, typically $k=4$, using fused image-text similarity.
GPT-4o then drafts captions, QA pairs, or diagnostic-workflow responses conditioned on the retrieved context, the source report fields, available ROI hints, and the study structure.
The prompt requires every clinical claim to be supported by either a source finding, a retrieved case, a guideline statement, an ROI, a view or slice reference, or a non-imaging clinical signal.
Unsupported details are not allowed to be introduced.

\paragraph{Unified instruction schema.}
The final instructions cover three formats.
Caption-style instructions ask for concise descriptions of visible findings, including anatomy, abnormality, laterality, and view or slice information when available.
VQA-style instructions pair an image or case context with a question and answer, optionally with a short evidence-based explanation.
Diagnosis-workflow instructions use case-level inputs, including multi-view images and clinical context, and produce evidence, image findings, differential discussion, diagnosis, and next step when available.
Each instance stores machine-readable provenance fields, including study id, modality, series or view, ROI references, clinical signals, findings, differentials, diagnosis, next step, and evidence citations.

\paragraph{Negative and normal controls.}
To reduce shortcut learning, we add normal controls, absent-lesion distractors, and near-miss distractors.
Normal controls require the model to justify the absence of acute findings.
Absent-lesion distractors are plausible diseases from guidelines or similar cases but are not supported by the current case.
Near-miss distractors refer to adjacent regions, alternate views, or similar-looking findings.
For each negative item, the target response must explain why the distractor is not supported, such as an incorrect region, missing view evidence, or contradictory clinical signal.

\paragraph{Physician validation of synthesized reasoning chains.}
Because part of MultiViewDx is constructed through retrieval-augmented synthesis, we validate generated reasoning chains with physician review.
We randomly sample 100 synthesized cases and ask three physicians with more than 10 years of clinical experience to independently review the original source case, associated imaging evidence, clinical context, and generated reasoning chain.
Each case is labeled as fully consistent, inconsistent but clinically correct, or incorrect or misleading.
A fully consistent case has reasoning and diagnosis supported by the source case and its evidence.
An inconsistent but clinically correct case reaches an acceptable final conclusion but has incomplete or imprecise intermediate reasoning or evidence attribution.
An incorrect or misleading case contains unsupported claims, contradicts source evidence, or could lead to an incorrect diagnostic interpretation.
Under this protocol, 88\% of sampled cases are fully consistent, 10\% are inconsistent but clinically correct, and 2\% are incorrect or misleading.
We report Fleiss' $\kappa=0.81$ for the three-way nominal labels.

\paragraph{Instruction formats.}
MultiViewDx contains three instruction formats. 
First, caption-style instructions use one image, one view, or a small set of views as input, and the target output is a concise description of visible findings, including anatomy, abnormality, laterality, and view or slice information when available.
Second, VQA-style instructions use an image or case context together with a natural-language question as input, and the target output is the answer to the question, optionally with a short evidence-based explanation when the source annotation supports it.
Third, diagnosis-workflow instructions use case-level inputs, including multi-view images and available clinical context, and the target output follows our evidence-linked workflow: evidence, image findings, discussion or differential diagnosis, final diagnosis, and when available, the next diagnostic or treatment-related step.
These formats allow the same dataset to supervise image understanding, medical QA, and case-level diagnostic reasoning.

\paragraph{Human validation of synthesized reasoning chains.}
Because part of MultiViewDx is constructed through retrieval-augmented synthesis, we conduct a physician validation study to assess the reliability of the generated reasoning chains.
We randomly sample 100 cases from the synthesized portion of MultiViewDx and ask three physicians, each with more than 10 years of clinical experience, to independently review them.
For each case, annotators are shown the original source case, the associated imaging evidence and clinical context, and the generated evidence-linked reasoning chain.
We use three outcome categories.
A case is marked as \emph{fully consistent} if the generated reasoning and diagnosis are supported by the source case and its evidence.
It is marked as \emph{inconsistent but clinically correct} if the final diagnostic conclusion is clinically acceptable but some intermediate reasoning step or evidence attribution is incomplete or imprecise.
It is marked as \emph{incorrect or misleading} if the output contains unsupported clinical claims, contradicts the source evidence, or could lead to an incorrect diagnostic interpretation.
Under this protocol, 88\% of sampled cases are judged fully consistent, 10\% are judged inconsistent but clinically correct, and 2\% are identified as incorrect or misleading.
We measure inter-annotator agreement using Fleiss' $\kappa$, treating the three validation labels as nominal categories, and obtain $\kappa=0.81$.
These results suggest that the synthesized reasoning chains are generally reliable, while also motivating our use of expert-authored cases in the final sequential fine-tuning stage.

\section{Training and Evaluation Pipeline}
\label{sec:training_eval_pipeline}

This section defines how MultiViewDx is used for training and evaluation.
We first describe the three training stages and the controlled variants used in ablations.
We then define the benchmark splits, baseline groups, and diagnostic-reasoning metrics.
All results are reported in Sec.~\ref{sec:results}.

\subsection{Training Pipeline}
\label{training}

We train MultiViewDx-8B in three stages.
Stage 1 performs medical image-text pretraining on 14M filtered PMC samples after removing corrupted items, non-medical examples, and captions with fewer than 20 words.
Only the adapter is updated, using learning rate $1\mathrm{e}{-4}$, image token length 512, DeepSpeed Stage 2, batch size 8, gradient accumulation 6, and 4 H100 GPUs for 420 hours.
Stage 2 performs instruction tuning on GPT-4o-synthesized MultiViewDx instructions, updating all parameters with learning rate $2\mathrm{e}{-5}$, DeepSpeed Stage 2, batch size 6, gradient accumulation 6, and 8 H100 GPUs for 80 hours.
Stage 3 sequentially fine-tunes the model on human-annotated clinical cases with learning rate $1\mathrm{e}{-5}$.
The full pipeline uses approximately 2,400 H100 GPU-hours and an estimated $0.81$--$1.0\times10^{21}$ FLOPs.

\begin{revaddblock}
\paragraph{Multi-image representation.}
MultiViewDx-8B is initialized from an LLaVA-8B model. Each image is independently encoded by the shared vision encoder and represented as a block of 512 visual tokens. For a multi-image case, visual blocks are serialized in case order with available modality, view, or slice identifiers in the accompanying text; standard one-dimensional sequence positions continue across the serialized visual and textual input. CT and MRI use the selected 2D slices described in Sec.~\ref{data_source_section}, ordered by slice position when available.
\end{revaddblock}

\paragraph{Training variants.}
We use six variants to isolate the data and training components.

\begin{itemize}
    \item \textbf{MultiViewDx-8B-20M} uses all available slices from 3D studies during training, and tests whether more visual context alone improves performance.
    \item \textbf{MultiViewDx-8B-Synthesis} \revmod{uses Stage~1 PMC image-text pretraining followed by GPT-4o-synthesized instruction tuning with the slice-capped case construction and stops before the human-annotated Stage~3.}
    \item \textbf{MultiViewDx-8B-Mix} combines GPT-4o-synthesized instructions and human-annotated clinical cases in a single SFT stage.
    \item \textbf{MultiViewDx-8B-AN} first trains on synthesized instructions and then sequentially fine-tunes on human-annotated clinical cases.
    \item \textbf{MultiViewDx-Non-Multi-View} keeps diagnostic-reasoning targets and human-annotated cases, but breaks case-level multi-view grouping into single-view or independently sampled examples.
    \item \textbf{MultiViewDx-NoReasoning} keeps case-level inputs and human-annotated data, but reduces target outputs to direct answers or diagnoses without the evidence-linked diagnostic trajectory.
\end{itemize}

\revmod{The last two variants test case-fragmented supervision and workflow-sparse supervision.}
\revadd{The Non-Multi-View ablation jointly changes case organization and the associated training distribution. Appendix~\ref{apx:paired_nomv} illustrates the evidence-linking contrast through paired same-case outputs and reports an inference-time view-removal test.}

\subsection{Evaluation Pipeline}
\label{sec:evaluation_pipeline}

\paragraph{Training and evaluation splits.}
For VQA-RAD, SLAKE, PathVQA, and PMC-VQA, we follow the official dataset partitions.
Only official training splits are used during fine-tuning, and all reported results use the corresponding official held-out test splits.
For the JAMA Clinical Challenge, we construct a disjoint case-level split because the dataset does not provide the same predefined train/test split as the VQA benchmarks: 60\% of cases are used for training and 40\% are reserved for testing.
No held-out MedVQA test instance or held-out JAMA case is used during training.
\revadd{MultiViewDx-8B-AN uses all four official benchmark training partitions.} \revadd{PMC-VQA$^{*}$ shares PMC/PMC-OA source provenance with the Stage~1 corpus.}

\paragraph{Evaluation axes.}
We evaluate two axes.
The first is \revmod{target-supervised performance on official held-out MedVQA test splits}, measured by accuracy on VQA-RAD~\cite{lau2018dataset}, SLAKE~\cite{liu2021slake}, PathVQA~\cite{he2020pathvqa}, and PMC-VQA~\cite{zhang2023pmc}.
The second is case-based diagnostic reasoning, measured on held-out JAMA Clinical Challenge cases that require integration of case context and imaging evidence.

For existing MedVQA benchmarks, we report answer accuracy on the official held-out test sets.
For JAMA, we report close-ended accuracy when applicable, UMLS-F1 factuality, GPT-4o rubric scores, and clinician ratings.
UMLS-F1 measures medical concept overlap using SciSpacy/UMLS, as described in Appendix~\ref{apx:umls-f}.
The GPT-4o rubric is a blinded 0--2 scoring protocol designed with clinicians.
It contains three axes: \textbf{Key Points}, which measures coverage of critical clinical elements; \textbf{Inference}, which measures the correctness and completeness of the diagnostic reasoning path; and \textbf{Evidence}, which measures whether claims are grounded in specific findings, imaging views or ROIs, and clinical signals.
Overall is the mean of the three rubric axes.
To reduce stylistic bias, outputs are style-normalized before judging, and the judge sees extracted gold summaries of Key Points, Inference, and Evidence rather than the full reference text.

\revmod{We also conduct a separate clinician evaluation on 100 randomly sampled held-out JAMA cases. Three clinicians review each case and its gold rationale and assign one holistic rating to the model rationale while considering Key Points, Inference, and Evidence. The averaged and normalized rating is \textbf{Doctor Score}, our primary human evaluation of model rationales.} \revadd{GPT-4o and DeepSeek provide complementary automatic scores; all three evaluations rank MultiViewDx-AN first.}

\paragraph{Knowledge and inference item classification.}
We estimate the proportion of knowledge-oriented and inference-oriented items using a fixed GPT-4o judging prompt.
For each benchmark item, the input to the judge includes the question, answer choices when available, and the reference answer.
The judge assigns one of two labels.
A question is labeled knowledge if it mainly requires recognition of a visual pattern, recall of a medical fact, or a short image-text association.
A question is labeled inference if it requires combining multiple findings, using clinical context, comparing possible diagnoses, reasoning across views or modalities, or making a multi-step clinical judgment.
We apply the same prompt to all datasets to characterize the proportion of knowledge-oriented and inference-oriented items, as shown in Appendix Fig.~\ref{fig:ratio}.
The full prompt is provided in Appendix~\ref{apx:knowledge_inference_prompt}.
These labels are used only for aggregate characterization of benchmark style, not as gold diagnostic labels for model evaluation.

\section{Results}
\label{sec:results}

We organize results around four questions.
First, does MultiViewDx retain strong \revmod{performance} on existing MedVQA benchmarks?
Second, does it improve case-based diagnostic reasoning beyond MedVQA accuracy?
Third, which components of the data design and training schedule matter?
Fourth, do clinician judgments and qualitative cases support the automatic results?
Additional baseline details, training configurations, and prompt templates are provided in Appendix~\ref{apx:experiment_details}.

\subsection{\revmod{Performance} on Existing MedVQA Benchmarks}
\label{sec:vqa_results}

Table~\ref{tab:result_tradition} compares MultiViewDx-8B-AN with proprietary models, general-purpose multimodal VLMs, and medical multimodal VLMs on VQA-RAD, SLAKE, PathVQA, and PMC-VQA$^{*}$.
MultiViewDx-8B-AN achieves the best average accuracy among the compared systems, reaching 79.0\%.
It outperforms the strongest medical baseline Hulu-Med-7B (74.3\%) and HuatuoGPT-Vision-34B (66.7\%), and also exceeds strong general-purpose models such as Gemini-2.5-Flash (63.8\%), GPT-4.1 (62.0\%), and Claude3-Opus (55.7\%).
\revmod{MultiViewDx-8B-AN is trained on all four official benchmark training partitions and evaluated on the corresponding held-out test sets. Appendix Tables~\ref{tab:exposure_table2}--\ref{tab:exposure_table3} report model-by-model training exposure.}

\noindent\textbf{Fine-tuning on benchmark training sets.} We additionally report a separate fine-tuning-on-training-set comparison in Table~\ref{tab:finetune_trainset}.
In this setting, each model is fine-tuned on the corresponding benchmark training split before evaluation on the held-out test split.
\revmod{Every row uses benchmark-specific supervised adaptation; Appendix Table~\ref{tab:exposure_table3} records the corresponding training exposure.}

\begin{table}[!ht]
\centering
\scriptsize
\setlength{\tabcolsep}{4.2pt}
\renewcommand{\arraystretch}{1}
\resizebox{\linewidth}{!}{
\begin{tabular}{lccccc}
\toprule
\textbf{Model} & \textbf{VQA-RAD} & \textbf{SLAKE} & \textbf{PathVQA} & \textbf{PMC-VQA\revmod{$^{*}$}} & \textbf{Avg.} \\
\midrule

\rowcolor{blue!12}
\multicolumn{6}{l}{\textbf{Proprietary Models}}\\
\rowcolor{blue!6} GPT-4.1            & 65.0 & 72.2 & 55.5 & 55.2 & 62.0 \\
\rowcolor{blue!6} GPT-4o             & 61.0 & 71.2 & 55.5 & 49.7 & 59.3 \\
\rowcolor{blue!6} Claude Sonnet 4    & 67.6 & 70.6 & 54.2 & 54.4 & 61.7 \\
\rowcolor{blue!6} Gemini-2.5-Flash   & 68.5 & 75.8 & 55.4 & 55.4 & 63.8 \\
\rowcolor{blue!6} GPT-4o-mini        & 45.9 & 59.0 & 37.9 & 33.3 & 44.0 \\
\rowcolor{blue!6} Claude3-Opus       & 52.5 & 55.2 & 54.3 & 60.7 & 55.7 \\
\midrule

\rowcolor{yellow!18}
\multicolumn{6}{l}{\textbf{General-purpose Multimodal VLMs}}\\
\rowcolor{yellow!8} Qwen2.5VL-7B       & 63.2 & 66.8 & 44.1 & 51.9 & 56.5 \\
\rowcolor{yellow!8} Janus-Pro-7B       & 49.7 & 55.2 & 35.4 & 50.1 & 47.6 \\
\rowcolor{yellow!8} InternVL2.5-8B     & 59.4 & 69.0 & 42.1 & 51.3 & 55.5 \\
\rowcolor{yellow!8} InternVL3-8B       & 65.4 & 72.8 & 48.6 & 53.8 & 60.1 \\
\rowcolor{yellow!8} Llama3.2-11B       & 58.8 & 65.8 & 32.9 & 48.1 & 51.4 \\
\rowcolor{yellow!8} InternVL3-14B      & 66.3 & 72.8 & 48.0 & 54.1 & 60.3 \\
\rowcolor{yellow!8} Qwen2.5VL-32B      & 71.8 & 71.2 & 41.9 & 54.5 & 59.9 \\
\rowcolor{yellow!8} InternVL2.5-38B    & 61.4 & 70.3 & 46.9 & 57.2 & 59.0 \\
\rowcolor{yellow!8} InternVL3-38B      & 65.4 & 72.7 & 51.0 & 56.6 & 61.4 \\
\rowcolor{yellow!8} Qwen-VL-Chat       & 47.0 & 56.0 & 55.1 & 36.6 & 48.9 \\
\rowcolor{yellow!8} Yi-VL-34B          & 53.0 & 58.9 & 47.3 & 39.5 & 49.7 \\
\rowcolor{yellow!8} LLaVA-7B           & 52.6 & 57.9 & 47.9 & 35.5 & 48.5 \\
\rowcolor{yellow!8} LLaVA-13B          & 55.8 & 58.9 & 51.9 & 36.6 & 50.8 \\
\rowcolor{yellow!8} LLaVA-34B          & 58.6 & 67.3 & 59.1 & 44.4 & 57.4 \\
\rowcolor{yellow!8} LLaVA-8B           & 54.2 & 59.4 & 54.1 & 36.4 & 51.0 \\
\midrule

\rowcolor{violet!16}
\multicolumn{6}{l}{\textbf{Medical Multimodal VLMs}}\\
\rowcolor{violet!7} BiomedGPT          & 16.6 & 13.6 & 11.3 & 27.6 & 17.3 \\
\rowcolor{violet!7} Med-R1-2B          & 39.0 & 54.5 & 15.3 & 47.4 & 39.0 \\
\rowcolor{violet!7} MedVLM-R1-2B       & 49.2 & 56.3 & 36.0 & 48.8 & 47.6 \\
\rowcolor{violet!7} HealthGPT-M3       & 56.8 & 70.8 & 55.4 & 55.4 & 59.6 \\
\rowcolor{violet!7} BioMedix2-8B       & 55.7 & 54.1 & 34.6 & 41.8 & 46.5 \\
\rowcolor{violet!7} MedGemma-4B-IT     & 72.3 & 78.2 & 48.1 & 49.2 & 62.0 \\
\rowcolor{violet!7} HuatuoGPT-V-7B     & 67.6 & 68.1 & 44.8 & 53.1 & 58.4 \\
\rowcolor{violet!7} Lingshu-7B         & 67.9 & 83.1 & 61.9 & 56.3 & 67.3 \\
\rowcolor{violet!7} Med-Flamingo       & 45.4 & 43.5 & 54.7 & 23.3 & 41.7 \\
\rowcolor{violet!7} RadFM              & 50.6 & 34.6 & 38.7 & 25.9 & 37.5 \\
\rowcolor{violet!7} LLaVA-Med-7B        & 51.4 & 48.6 & 56.8 & 24.7 & 45.4 \\
\rowcolor{violet!7} + PubMedVision       & 63.8 & 74.5 & 59.9 & 52.7 & 62.7 \\
\rowcolor{violet!7} HuatuoGPT-Vision-34B & 68.1 & 76.9 & 63.5 & 58.2 & 66.7 \\
\rowcolor{violet!7} Hulu-Med-7B        & 78.0 & 86.8 & 65.6 & \revmod{\textbf{66.8}} & 74.3 \\
\rowcolor{gray!20}
\multicolumn{6}{l}{\textbf{Our Model}}\\
\rowcolor{gray!7} MultiViewDx-8B-AN    & \textbf{86.1} & \textbf{87.7} & \textbf{80.4} & \revmod{61.9} & \textbf{79.0} \\

\bottomrule
\end{tabular}
}

\caption{
\revmod{Main comparison on official held-out medical VQA test splits. MultiViewDx-8B-AN uses the four official training partitions. Additional MultiViewDx variants appear in Table~\ref{tab:vqa_ablation}, and Appendix Table~\ref{tab:exposure_table2} reports exposure for every compared model. $^{*}$PMC-VQA and Stage~1 share PMC/PMC-OA source provenance.}
}
\label{tab:result_tradition}
\end{table}

\begin{table}[t]
\centering
\scriptsize
\setlength{\tabcolsep}{3pt}
\renewcommand{\arraystretch}{1.05}
\resizebox{\linewidth}{!}{
\begin{tabular}{lccccc}
\toprule
\textbf{Model} & \textbf{VQA-RAD} & \textbf{SLAKE} & \textbf{PathVQA} & \textbf{PMC-VQA\revmod{$^{*}$}} & \textbf{Avg.} \\
\midrule
LLAVA-v1.5-LLAMA3-8B & 63.3 & 68.9 & 85.2 & 50.3 & 66.9 \\
LLAVA\_Med-8B & 66.3 & 69.5 & 90.7 & 52.7 & 69.8 \\
HuatuoGPT-Vision-8B & 68.9 & 84.1 & \revmod{\textbf{93.0}} & 57.3 & 75.8 \\
MultiViewDx & \textbf{88.3} & \textbf{91.1} & \revmod{92.7} & \textbf{88.6} & \textbf{90.2} \\
\bottomrule
\end{tabular}
}
\caption{
\revmod{Benchmark-specific supervised adaptation results. Every row is fine-tuned on the corresponding official training partition and evaluated on the held-out test split. Appendix Table~\ref{tab:exposure_table3} summarizes the setting. $^{*}$PMC-VQA shares the source provenance noted in Table~\ref{tab:result_tradition}.}
}
\label{tab:finetune_trainset}
\end{table}

\subsection{Case-Based Diagnostic Reasoning Beyond MedVQA}
\label{sec:jama_results}

Existing MedVQA benchmarks test important medical image QA skills, but they do not fully test whether a model can integrate imaging evidence with patient context and justify a diagnostic decision.
We therefore evaluate on held-out JAMA Clinical Challenge cases, which provide real-world clinical vignettes with case context and associated visual evidence.
Table~\ref{jama_result} reports close-ended accuracy, UMLS-F1 factuality, GPT-4o rubric scores for Key Points, Inference, and Evidence when available.

MultiViewDx-AN achieves the best Accuracy (58.5\%), the highest UMLS Factuality (0.23), and the best GPT-4 Overall score ($1.29\pm0.02$). It also obtains the highest Key-Points score (1.28), strong Inference (1.32), and high Evidence (1.27). Among baselines, Claude3-Opus achieves the strongest Inference score (1.56) but weaker Evidence (0.67), while Huatuo-34B provides the strongest Evidence score among baselines (1.31) but lower Inference (1.06). Hulu-Med-7B and Lingshu-7B obtain relatively strong overall rubric scores ($1.22\pm0.03$ and $1.18\pm0.04$), suggesting that medical-domain training improves evidence-oriented outputs, but still does not fully match MultiViewDx-AN on overall case-based diagnostic reasoning.

Fig.~\ref{fig:ratio} further shows that JAMA contains a larger share of inference-oriented items than existing MedVQA benchmarks.
We use it as a complementary diagnosis-oriented evaluation, not as a replacement for MedVQA accuracy.

\begin{table}[t]
\centering
\scriptsize
\setlength{\tabcolsep}{3.2pt}
\renewcommand{\arraystretch}{1.03}
\begin{tabular}{lcccccc}
\toprule
\textbf{Model} & \textbf{Acc.} & \textbf{UMLS} & \textbf{KP} & \textbf{Inf.} & \textbf{Evi.} & \textbf{Overall} \\
\midrule
\textbf{C3-Opus}  & 58.4 & 0.18 & 1.27 & \textbf{1.56} & 0.67 & 1.17 $\pm$ 0.04 \\
\textbf{G4o-mini} & 46.2 & 0.16 & 0.99 & 1.13 & 0.60 & 0.91 $\pm$ 0.06 \\
\textbf{H-7B}     & 34.5 & 0.13 & 1.11 & 1.06 & 1.08 & 1.08 $\pm$ 0.03 \\
\textbf{H-34B}    & 44.7 & 0.16 & 1.01 & 1.06 & \textbf{1.31} & 1.13 $\pm$ 0.05 \\
\textbf{MG-4B}    & 45.8 & 0.19 & 1.06 & 1.10 & 1.21 & 1.14 $\pm$ 0.03 \\
\textbf{LS-7B}    & 53.6 & 0.19 & 1.10 & 1.19 & 1.27 & 1.18 $\pm$ 0.04 \\
\textbf{HM-7B}    & 54.6 & 0.18 & 1.15 & 1.22 & 1.26 & 1.22 $\pm$ 0.03 \\
\textbf{MVDX-AN}  & \textbf{58.5} & \textbf{0.23} & \textbf{1.28} & 1.32 & 1.27 & \textbf{1.29 $\pm$ 0.02} \\
\bottomrule
\end{tabular}
\caption{
Results on the JAMA Clinical Challenge.
Acc.=close-ended accuracy; UMLS=UMLS-F1 factuality; KP=Key Points; Inf.=Inference; Evi.=Evidence.
Model abbreviations: C3-Opus=Claude3-Opus, G4o-mini=GPT-4o-mini, H-7B/H-34B=Huatuo-7B/34B, MG-4B=Med-Gemma-4B-IT, LS-7B=Lingshu-7B, HM-7B=Hulu-Med-7B, and MVDX-AN=MultiViewDx-AN.
We additionally rescore a subset with DeepSeek in Appendix Table~\ref{tab:model_comparison}; the relative ranking is consistent.
}
\label{jama_result}
\end{table}

\noindent\textbf{Why fixing Case-fragmented supervision helps and why traditional metrics can mislead.} Traditional VQA sets are dominated by \emph{knowledge} items (SLAKE 78.1\%, VQA-RAD 76.4\%, PathVQA 69.2\%), so models can score well by recalling single-view facts or template associations; JAMA contains far fewer knowledge items (44.9\%) and instead requires \emph{inference} from \emph{multiple} views/modalities (Fig.~\ref{fig:ratio}).
\revmod{This mismatch explains why surface accuracy on VQA can overstate clinical readiness: those tests rarely require cross-view corroboration, ROI grounding, or contradiction checks. Case-level supervision supports triangulation across views, disambiguation through view-dependent cues, evidence attribution to specific views or ROIs, and study-level coherence. All values reported here match Table~\ref{jama_result}; Appendix Table~\ref{tab:model_comparison} provides the DeepSeek sensitivity analysis.}
\revadd{On 100 multi-view cases, removing the view containing the relevant image evidence while retaining the vignette and remaining inputs reduced accuracy to 31\%. The result shows that view-specific evidence is essential during inference.}

\subsection{Ablations: Workflow, Case Organization, and Training Schedule}
\label{sec:ablation}


\begin{table}[t]
\centering
\scriptsize
\setlength{\tabcolsep}{3pt}
\renewcommand{\arraystretch}{1.05}
\resizebox{\linewidth}{!}{
\begin{tabular}{lccccc}
\toprule
\textbf{Variant} & \textbf{VQA-RAD} & \textbf{SLAKE} & \textbf{PathVQA} & \textbf{PMC-VQA\revmod{$^{*}$}} & \textbf{Avg.} \\
\midrule
MultiViewDx-8B-20M & 67.8 & 76.1 & 57.8 & 53.6 & 63.8 \\
MultiViewDx-8B-Synthesis & 69.2 & 77.2 & 63.6 & 58.4 & 67.1 \\
MultiViewDx-8B-Mix & 74.2 & 81.3 & 76.3 & 59.1 & 72.2 \\
MultiViewDx-8B-AN & \textbf{86.1} & \textbf{87.7} & \textbf{80.4} & \textbf{61.9} & \textbf{79.0} \\
\bottomrule
\end{tabular}
}
\caption{
Ablation of MultiViewDx training variants on standard medical VQA benchmarks.
MultiViewDx-8B-20M uses all available 3D slices during training;
Synthesis uses GPT-4o-synthesized instruction data with slice-capped construction;
Mix combines synthesized and human-annotated cases in one SFT stage;
AN first trains on synthesized data and then sequentially fine-tunes on human-annotated cases. \revmod{$^{*}$PMC-VQA shares the source provenance noted in Table~\ref{tab:result_tradition}.}
}
\label{tab:vqa_ablation}
\end{table}

We use the variants defined in Sec.~\ref{training} to test which parts of MultiViewDx matter.
The ablations are designed to answer four questions.
First, does a compact slice-capped case representation work better than feeding all available 3D slices?
Second, is sequential fine-tuning on human-annotated clinical cases more effective than mixing synthetic and human cases in one SFT stage?
Third, does case-level multi-view organization matter after controlling for reasoning supervision and human cases?
Fourth, does explicit evidence-linked reasoning supervision matter after controlling for multi-view inputs and human cases?

\paragraph{Training schedule ablation.}
Table~\ref{tab:vqa_ablation} shows the effect of slice selection and training schedule on standard medical VQA benchmarks. MultiViewDx-8B-20M, which uses all available slices from 3D studies, obtains 63.8 average accuracy. Using slice-capped GPT-4o-synthesized instruction data improves the average score to 67.1, suggesting that densely adjacent slices do not necessarily help when they introduce redundant visual context. Mixing synthesized data with human-annotated clinical cases further improves the score to 72.2. The best result is obtained by MultiViewDx-8B-AN, which first trains on synthesized instruction data and then sequentially fine-tunes on human-annotated cases, reaching 79.0 average accuracy. This suggests that human-annotated cases are most effective when used as a later expert-case alignment stage rather than simply mixed into one SFT stage.

\paragraph{Component-level clinician evaluation.}
Table~\ref{tab:component_doctor} complements the automatic VQA ablation with clinician evaluation on JAMA. MultiViewDx-Synthesis uses synthetic instruction data and diagnosis-reasoning supervision but does not include case-level multi-view grouping or human-annotated case fine-tuning, obtaining a Doctor Score of 0.30. MultiViewDx-Non-Multi-View adds human-annotated cases but removes case-level multi-view organization, yielding a Doctor Score of 0.29. MultiViewDx-NoReasoning keeps multi-view inputs and human-annotated cases but removes explicit diagnosis-reasoning supervision, leading to the lowest Doctor Score of 0.11. MultiViewDx-AN includes all components and obtains the highest Doctor Score of 0.53. \revmod{The NoMV--AN gap shows the value of case-level organization under the associated training distribution.} \revadd{Appendix Table~\ref{tab:paired_nomv_case} illustrates the evidence-linking contrast on the same case, and Appendix Table~\ref{tab:view_removal} reports the 100-case view-removal result.}

\begin{table}[t]
\centering
\scriptsize
\setlength{\tabcolsep}{3.4pt}
\renewcommand{\arraystretch}{1.03}
\begin{tabular}{lccccc}
\toprule
\textbf{Model} & \textbf{Syn.} & \textbf{MV} & \textbf{Reason.} & \textbf{HA} & \textbf{Doc.} \\
\midrule
\textbf{C3-Opus} & - & - & - & - & 0.48 \\
\textbf{H-7B}    & \checkmark & $\times$ & $\times$ & $\times$ & 0.21 \\
\textbf{H-34B}   & \checkmark & $\times$ & $\times$ & $\times$ & 0.35 \\
\textbf{MG-4B}   & \checkmark & $\times$ & $\times$ & $\times$ & 0.40 \\
\textbf{LS-7B}   & \checkmark & $\times$ & \checkmark & $\times$ & 0.44 \\
\textbf{HM-7B}   & \checkmark & $\times$ & \checkmark & $\times$ & 0.50 \\
\midrule
\textbf{MVDX-Syn}  & \checkmark & $\times$ & \checkmark & $\times$ & 0.30 \\
\textbf{MVDX-NoMV} & \checkmark & $\times$ & \checkmark & \checkmark & 0.29 \\
\textbf{MVDX-NoR}  & \checkmark & \checkmark & $\times$ & \checkmark & 0.11 \\
\textbf{MVDX-AN}   & \checkmark & \checkmark & \checkmark & \checkmark & \textbf{0.53} \\
\bottomrule
\end{tabular}
\caption{\revmod{Component-level clinician evaluation on 100 held-out JAMA cases. Doctor Score (Doc.) is the averaged and normalized holistic rating from three clinicians, who consider Key Points, Inference, and Evidence with reference to the case and gold rationale. Syn.=synthetic or teacher-distilled instruction data; MV=case-level multi-view grouping; Reason.=evidence-linked diagnostic reasoning targets; HA=human-annotated clinical cases. Baseline checkmarks denote analogous training signals reported in the original papers.}}
\label{tab:component_doctor}
\end{table}

\subsection{Human Evaluation and Qualitative Analysis}
\label{sec:human_eval_qualitative}

The clinician evaluation in Table~\ref{tab:component_doctor} assesses whether generated rationales are clinically acceptable beyond final-answer matching.
MultiViewDx-AN obtains the highest Doctor Score (0.53), above Hulu-Med-7B (0.50) and Claude3-Opus (0.48).
The margin is modest, so we interpret it together with the ablations and qualitative cases rather than treating the score alone as sufficient evidence.

The qualitative JAMA case in Appendix Table~\ref{tab:case1} illustrates the same pattern. Appendix Table~\ref{tab:paired_nomv_case} additionally contrasts illustrative Full and Non-Multi-View outputs on the same case.
The case requires linking bilateral optic nerve edema and macular edema with pregnancy, hypertension, and chronic kidney disease, then prioritizing blood pressure assessment to rule out malignant hypertension.
MultiViewDx connects the visual findings with the patient context and explains why vital-sign assessment is the immediate next step.
Claude3-Opus selects the correct action but gives a less evidence-specific rationale, while HuatuoGPT-Vision variants focus on diabetes-related testing and miss the urgent prioritization.
The example suggests that MultiViewDx's advantage lies in evidence grounding and clinical prioritization, not only in final answer selection.

\subsection{Interpretation}
\label{sec:ws_cf_discussion}

The results support the two data-design claims in different ways.
MedVQA gains show that workflow-aware, report-anchored supervision improves \revmod{medical image QA performance on held-out test sets}.
The JAMA results and ablations provide the stronger test of diagnostic reasoning: removing case-level multi-view grouping weakens performance even when reasoning supervision and human cases remain, and removing evidence-linked reasoning targets causes the largest drop in clinician-rated quality.
Thus, the main effect is not explained by parameter scale or by adding more images alone; it comes from organizing supervision around evidence-linked clinical cases.

\section{Limitations and Ethical Considerations}

MultiViewDx has several limitations.
First, modality and anatomy coverage remain uneven because high-quality human-validated clinical data are scarce.
Performance may therefore vary across body regions, acquisition protocols, institutions, and patient populations that are underrepresented in the corpus.
Second, our representation of 3D CT and MRI studies uses selected 2D slices rather than full volumetric reasoning.
The slice-capped strategy is practical and empirically useful in our experiments, but it does not replace architectures designed for full 3D fusion.
Third, part of the corpus is retrieval-synthesized.
Although physician review finds a low rate of incorrect or misleading reasoning in the sampled cases, this does not eliminate all unsupported statements or subtle evidence-attribution errors.
Fourth, clinician evaluation is conducted on a finite sample of held-out JAMA cases and should be expanded across specialties and institutions.
\revadd{Current validation includes a random, unstratified physician audit of 100 synthesized cases. Future audits will stratify by source, modality, diagnosis, and synthesis workflow. PMC-VQA and Stage~1 share PMC/PMC-OA provenance, and the retained PMC manifests provide sample-level counts rather than unique patient-, study-, or case-level counts. Future validation will quantify retrieval relevance and atomic claim entailment, clinically review the 300,000-instance Chinese subset, and calibrate automatic rubric scores against clinician ratings on matched outputs.}

The dataset and model are intended for research on medical multimodal reasoning, not for autonomous diagnosis.
Any clinical use would require prospective validation, institution-specific calibration, uncertainty reporting, bias and subgroup analysis, and human oversight.
We also avoid using identifiable patient information and preserve only de-identified clinical context, but de-identification and data governance should be re-audited before any redistribution or deployment.

\section{Conclusion}

We identify two data limitations that make medical MLLMs appear stronger on MedVQA than on diagnosis-oriented cases: workflow-sparse supervision and case-fragmented supervision.
MultiViewDx addresses these limitations by using evidence-linked clinical cases as the training unit, aligning multi-view imaging evidence with patient context and staged diagnostic fields.
Fine-tuning on MultiViewDx improves \revmod{performance on existing MedVQA held-out test sets} and yields stronger performance on held-out JAMA case-based diagnostic reasoning.
Controlled ablations and clinician evaluation further show that the gains depend on both case-level multi-view organization and explicit evidence-linked reasoning targets.

\section*{Acknowledgments}
We thank the anonymous reviewers and Action Editor Mihai Surdeanu for their constructive feedback.

\bibliography{tacl2021}
\bibliographystyle{acl_natbib}
\newpage
\appendix

\section{Evaluation Metrics and Dataset Statistics}

\revmod{Doctor Score is the primary human evaluation of model rationales. GPT-4o and DeepSeek provide complementary automatic analyses, and MultiViewDx-AN ranks highest under all three evaluation routes among the displayed systems. Style normalization reduces surface-form differences before automatic judging. Table~\ref{tab:model_comparison} uses the DeepSeek score to compare model ordering on its own scale.}

\subsection{UMLS-F1 Factuality Metric}
\label{apx:umls-f}
To evaluate the factual accuracy of LLM outputs, we leverage the UMLS concept overlap metric. The Unified Medical Language System (UMLS)~\cite{bodenreider2004unified} enhances biomedical interoperability by unifying a comprehensive collection of biomedical terminologies, classification systems, and coding standards. By reconciling semantic variances and representational disparities across different biomedical repositories, UMLS facilitates standardized understanding.

We employ the Scispacy library\footnote{Using the Scispacy \textit{en\_core\_sci\_lg} model} to identify and align medical named entities in texts with their corresponding UMLS concepts. Scispacy excels in entity recognition, enabling accurate association of named entities in LLM outputs with relevant UMLS concepts, a critical capability for assessing factual accuracy.

Our analytical process utilizes precision and recall metrics. Precision measures the proportion of shared concepts between the LLM output and the ground truth, indicating factual correctness. Recall assesses how well the LLM output covers the concepts present in the ground truth, reflecting the relevance of the information. Formally, given the concept sets from the ground truth ($C_{\text{ref}}$) and the LLM output ($C_{\text{gen}}$), precision and recall are calculated as:

\begin{equation}
\text{Precision} = \frac{|C_{\text{ref}} \cap C_{\text{gen}}|}{|C_{\text{gen}}|},
\end{equation}

\begin{equation}
\text{Recall} = \frac{|C_{\text{ref}} \cap C_{\text{gen}}|}{|C_{\text{ref}}|}.
\end{equation}

The F1 score, derived from these precision and recall values, provides a balanced measure of the LLM output's accuracy and relevance.

\begin{table}[t]
\centering
\scalebox{0.65}{
\begin{tabular}{lcccc}
\toprule
\textbf{Model} & \textbf{Acc.} & \textbf{UMLS-F} & \textbf{GPT4-O} & \textbf{DS-O} \\
\midrule
Claude3-Opus      & 58.4 & 0.18 & 1.17 & 2.31 \\
GPT-4o-mini       & 46.2 & 0.16 & 0.91 & 1.95 \\
Huatuo-7B         & 34.5 & 0.13 & 1.08 & 2.06 \\
Huatuo-34B        & 44.7 & 0.16 & 1.13 & 2.24 \\
MultiViewDx       & 41.2 & 0.11 & 0.78 & 1.86 \\
MultiViewDx-AN    & \textbf{58.5} & \textbf{0.23} & \textbf{1.29} & \textbf{2.55} \\
\bottomrule
\end{tabular}
}
\caption{\revmod{Auxiliary automatic-evaluator sensitivity analysis. Acc.=Accuracy; UMLS-F=UMLS-based factuality; GPT4-O=GPT-4o Overall; DS-O=DeepSeek Overall. MultiViewDx-AN ranks highest under both automatic evaluators among the displayed rows; DS-O is interpreted on the DeepSeek scale.}}
\label{tab:model_comparison}
\end{table}

\subsection{JAMA Specialty Distribution}
\label{apx:jama_distribution}

\begin{table}[!ht]
\centering
\scriptsize
\setlength{\tabcolsep}{3pt}
\renewcommand{\arraystretch}{0.92}
\begin{tabular}{l r}
\toprule
 & \textbf{Count} \\
\midrule
Oto-HNS   & 513 \\
Ophth     & 466 \\
Derm      & 368 \\
JAMA      & 328 \\
JN Learn  & 299 \\
Surg      & 133 \\
Oncol     & 105 \\
Cardiol   & 92 \\
Neurol    & 61 \\
Pediatr   & 60 \\
Psych     & 6 \\
\bottomrule
\end{tabular}
\caption{
Distribution across JAMA specialty journals.
Abbr.: Oto-HNS=JAMA Otolaryngology--Head \& Neck Surgery, Ophth=JAMA Ophthalmology, Derm=JAMA Dermatology, Surg=JAMA Surgery, Oncol=JAMA Oncology, Cardiol=JAMA Cardiology, Neurol=JAMA Neurology, Pediatr=JAMA Pediatrics, Psych=JAMA Psychiatry, JN Learn=JN Learning.
}
\label{tab:jama_distribution}
\end{table}

\begin{figure}
    \centering
\includegraphics[width=0.8\linewidth]{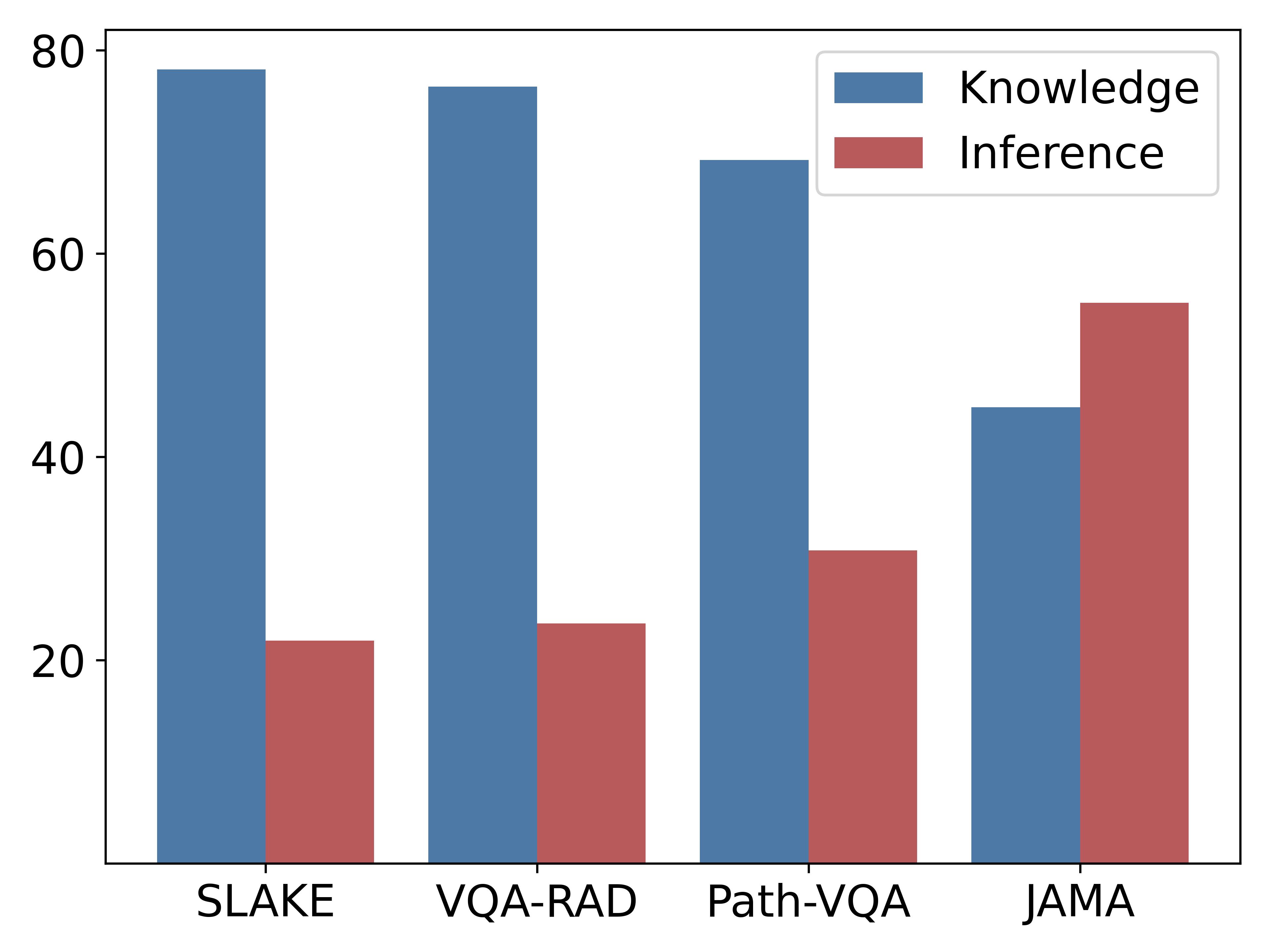}

        \caption{This figure illustrates the proportion of questions assessing knowledge and inference in the Slake, VQA-RAD, Path-VQA, and JAMA Clinical Challenge datasets.}
        \label{fig:ratio}

\end{figure}

\section{Experiment Settings and Data Construction Examples}
\label{apx:qualitative_examples}

\subsection{Experiment and Implementation Details} \label{apx:experiment_details} This section provides the experimental details that are summarized in Sec.~\ref{sec:training_eval_pipeline}. The main text focuses on the training and evaluation pipeline in Fig.~\ref{fig:pipeline}; here we provide the full baseline list, benchmark split details, training configuration, and variant definitions used in the results. \subsection{Baseline Models} \label{apx:baseline_models} We organize baselines into three groups: proprietary multimodal models, general-purpose open multimodal VLMs, and medical multimodal VLMs. This grouping matches the main benchmark comparison in Table~\ref{tab:result_tradition}. \paragraph{Proprietary models.} We include strong closed-source baselines that are commonly used as upper bounds for general multimodal reasoning, including GPT-4.1, GPT-4o, GPT-4o-mini, Claude3-Opus, Claude Sonnet 4, and Gemini-2.5-Flash. \paragraph{General-purpose multimodal VLMs.} We benchmark a diverse set of open VLMs across model families and scales, including Qwen2.5VL-7B/32B~\citep{bai2025qwen2}, Janus-Pro-7B~\citep{chen2025janus}, InternVL2.5-8B/38B~\citep{chen2024expanding}, InternVL3-8B/14B/38B~\citep{zhu2025internvl3}, Llama3.2-11B~\citep{meta2024llama}, Qwen-VL-Chat, Yi-VL-34B~\citep{young2024yi}, and LLaVA variants including LLaVA-7B/8B/13B/34B~\citep{liu2024llava}. \paragraph{Medical multimodal VLMs.} We compare against representative medical-domain models, including BiomedGPT~\citep{luo2023biomedgpt}, Med-R1-2B~\citep{lai2025med}, MedVLM-R1-2B~\citep{pan2025medvlm}, HealthGPT-M3~\citep{lin2025healthgpt}, BioMedix2-8B~\citep{mullappilly2024bimedix2}, MedGemma-4B-IT~\citep{sellergren2025medgemma}, HuatuoGPT-Vision-7B/34B~\citep{chen2024huatuogpt}, Lingshu-7B~\citep{xu2025lingshu}, Hulu-Med-7B~\citep{jiang2025hulu}, Med-Flamingo~\citep{moor2023med}, RadFM~\citep{wu2025towards}, LLaVA-Med~\citep{li2023llava}, and LLaVA-Med variants trained with PubMedVision-style data.

\subsection{Benchmark Splits} 
\label{apx:benchmark_splits} 

For VQA-RAD, SLAKE, PathVQA, and PMC-VQA, we follow the official dataset partitions. Only the official training split of each dataset is used during fine-tuning, and evaluation is conducted exclusively on the corresponding official held-out test split. Thus, although these datasets contribute training samples to MultiViewDx fine-tuning, their test samples are not used during training. For the JAMA Clinical Challenge, which does not provide the same standard train/test split as the VQA benchmarks, we construct a disjoint case-level split. We use 60\% of cases for training and reserve the remaining 40\% exclusively for testing. All JAMA accuracy, UMLS-F1, GPT-4o rubric, and Doctor Score results are reported on this held-out 40\% test split. \revadd{PMC-VQA$^{*}$ follows its official QA split and shares PMC/PMC-OA source provenance with Stage~1.} 

\subsection{Training Configuration and Compute} 

\label{apx:training_compute} 

Table~\ref{tab:training_config_appendix} summarizes the training configuration used for MultiViewDx-8B. The compute values are approximate and are reported to clarify the scale of training. 

\begin{table*}[t] \centering \scriptsize \setlength{\tabcolsep}{4pt} \renewcommand{\arraystretch}{1.08} \resizebox{\linewidth}{!}{ \begin{tabular}{lcccccc} \toprule \textbf{Stage} & \textbf{Data} & \textbf{Samples} & \textbf{LR} & \textbf{Updated Params.} & \textbf{GPU Setting} & \textbf{Approx. Compute} \\ \midrule Stage 1: Medical pretraining & Filtered PMC image-text data & 14M & $1\mathrm{e}{-4}$ & Adapter only & 4 H100 GPUs & $\sim$1,600 H100 hours \\ Stage 2: Instruction tuning & GPT-4o-synthesized MultiViewDx data & 4M & $2\mathrm{e}{-5}$ & Full model & 8 H100 GPUs & $\sim$640 H100 hours \\ Stage 3: Sequential fine-tuning & Human-annotated clinical cases & 1M & $1\mathrm{e}{-5}$ & Full model & 8 H100 GPUs & $\sim$160 H100 hours \\ \midrule \textbf{Total} & -- & -- & -- & -- & -- & $\sim$2,400 H100 hours \\ \bottomrule \end{tabular} } \caption{ Training configuration and approximate compute cost for MultiViewDx-8B. Stage 1 provides broad medical image-text grounding; Stage 2 adds instruction following and evidence-linked QA from GPT-4o-synthesized data; Stage 3 sequentially fine-tunes on human-annotated clinical cases. The overall estimated compute is on the order of $0.81$--$1.0\times10^{21}$ FLOPs. } \label{tab:training_config_appendix} \end{table*}

We adopt the three-stage training pipeline shown in Fig.~\ref{fig:pipeline}.
For pre-training~\cite{radford2018improving,chen2023meditron}, we filter the original PMC data by removing captions with fewer than 20 words, yielding 14M samples, and train with a learning rate $1\mathrm{e}{-4}$, image token length 512, DeepSpeed Stage 2 (batch size 8, gradient accumulation 6), updating only the adapter. This stage runs on 4 H100 GPUs for 420 hours.
For instruction tuning~\cite{longpre2023flan,wang2023self,tran2024bioinstruct,han2023medalpaca}, we fine-tune on MultiViewDx with learning rate $2\mathrm{e}{-5}$ under DeepSpeed Stage 2 (batch size 6, gradient accumulation 6), updating full parameters, and train on 8 H100 GPUs for 100 hours.
Overall, the full training pipeline uses approximately 2,400 H100 GPU-hours and has an estimated compute cost on the order of $0.81$--$1.0\times10^{21}$ FLOPs.
Finally, we apply the third stage in Fig.~\ref{fig:pipeline}: sequential fine-tuning on human-annotated clinical cases with learning rate $1\mathrm{e}{-5}$, using additional GPT-4o augmentation described in Appendix~\ref{template}.
This sequential fine-tuning stage is designed to further align the model with expert-authored diagnostic workflow, fine-grained evidence grounding, and case-level multi-view reasoning.

\subsection{Prompt for Knowledge/Inference Classification}
\label{apx:knowledge_inference_prompt}

We use GPT-4o to classify each benchmark item as knowledge-oriented or inference-oriented.

\begin{quote}
You are classifying medical QA examples into two categories.

Input:
Question: [QUESTION]

Answer choices: [CHOICES, if available]

Reference answer: [REFERENCE ANSWER]

Label the example as one of the following:

Knowledge: the question can be answered mainly by recognizing a visual pattern, recalling a medical fact, or matching a short image-text association. It does not require combining multiple evidence sources or multi-step clinical reasoning.

Inference: the question requires combining multiple findings, using clinical context, comparing differential diagnoses, reasoning across views or modalities, or making a multi-step clinical judgment.

Return only one label: Knowledge or Inference.
\end{quote}

\subsection{Template Prompt}
\label{template}

\noindent \textbf{Generate Instruction Data.} We construct instruction data with both closed-ended and open-ended formats. For close-ended datasets (e.g., PMC-VQA, AMBOSS, JAMA train, SLAKE train, VQA-RAD train, PathVQA), we generate only answer options. For open-ended tasks (mainly from JAMA), we additionally require step-by-step reasoning. We use GPT-4o to generate question-answer pairs from each image and its augmented caption (3-10 QAPs per caption), and we constrain generation to be grounded in the provided caption and image evidence (including ROI-level cues when available). Full prompt templates are shown in Github.

\noindent \textbf{Evaluation Pipeline Prompt.} For close-ended QA we report accuracy. For open-ended diagnostic reasoning, we use two stricter metrics: UMLS-based factuality and a GPT-4o rubric along \textbf{Key Points}, \textbf{Inference}, and \textbf{Evidence}. To reduce style bias, we first normalize model outputs into a consistent style, then ask the judge to score each aspect independently based on extracted summaries. Prompt details are provided in our Github.

\subsection{JAMA Case Study: Clinical Prioritization}
\label{apx:jama_case_study}

Table~\ref{tab:case1} presents a qualitative JAMA case study used to analyze diagnostic reasoning behavior across models.
The case requires more than recognizing ophthalmic image findings: the model must combine bilateral optic nerve edema, pregnancy, hypertension, and chronic kidney disease to prioritize immediate vital-sign assessment for possible malignant hypertension.
This example supports the main-paper discussion that clinical diagnostic reasoning is not fully captured by answer accuracy alone and that models may fail by over-weighting a salient but less urgent disease prior, such as diabetes, instead of selecting the clinically urgent next step.

\subsection{\revadd{Full-versus-Non-Multi-View Ablation Case}}
\label{apx:paired_nomv}
\revadd{Table~\ref{tab:paired_nomv_case} presents illustrative simulated outputs on the same JAMA Clinical Challenge case in Table~\ref{tab:case1}.}

\begin{table*}[p]
\centering
\small
\begin{tabularx}{0.98\textwidth}{>{\revaddcolor\bfseries\raggedright\arraybackslash}p{0.18\textwidth}>{\revaddcolor\raggedright\arraybackslash}X>{\revaddcolor\raggedright\arraybackslash}p{0.25\textwidth}}
\toprule
Model & Simulated generated output on the same input & Observable result\\
\midrule
MultiViewDx-AN (full multi-view) &
\textbf{Answer: C. Measure heart rate, respiratory rate, and blood pressure.} The right- and left-eye fundus photographs show bilateral optic-disc edema, more pronounced in the left eye with associated nerve-fiber-layer hemorrhage. OCT additionally shows center-involving macular edema in both eyes. In a 15-week pregnant patient with hypertension and chronic kidney disease, this bilateral fundus-plus-OCT pattern is concerning for hypertensive end-organ injury rather than isolated progression of diabetic retinopathy. The immediate next step is to check vital signs and blood pressure to evaluate for malignant hypertension with papillopathy. &
Correct. Explicitly links both fundus views, bilateral OCT findings, and systemic risk factors before selecting the urgent next step.\\
\midrule
MultiViewDx-NoMV (non-multi-view) &
\textbf{Answer: B. Determine blood glucose level and perform glycated hemoglobin testing.} The patient has type 1 diabetes and proliferative diabetic retinopathy. The retinal hemorrhages and bilateral macular edema are compatible with worsening diabetic retinopathy during pregnancy. Blood glucose and HbA1c should therefore be assessed to determine whether poor glycemic control is causing the visual decline and to guide diabetes management. &
Incorrect. Recognizes plausible local abnormalities but anchors on diabetes and fails to connect the bilateral optic-disc pattern and OCT findings with hypertension, kidney disease, and pregnancy.\\
\bottomrule
\end{tabularx}
\caption{\revadd{Illustrative simulated outputs from the full and Non-Multi-View settings on the JAMA case in Table~\ref{tab:case1}.}}
\label{tab:paired_nomv_case}
\end{table*}

\revadd{The full-model response connects bilateral fundus findings, OCT evidence, and systemic risk factors before selecting the urgent next step. The Non-Multi-View response recognizes local retinal abnormalities but anchors on diabetes, omits the cross-view clinical link, and selects the wrong action.}

\begin{table*}[p]
\centering
\small
\begin{tabular}{>{\revaddcolor}l>{\revaddcolor}c>{\revaddcolor}c}
\toprule
Inference condition & Cases & Accuracy\\
\midrule
Evidence-bearing view removed; vignette and remaining inputs retained & 100 & 31\%\\
\bottomrule
\end{tabular}
\caption{\revadd{Inference-time removal of the evidence-bearing view on 100 multi-view cases. Accuracy decreases to 31\%.}}
\label{tab:view_removal}
\end{table*}

\subsection{\revadd{Benchmark-Training Exposure Matrix}}
\label{apx:exposure_matrix}
\revadd{Table~\ref{tab:exposure_table2} records documented benchmark-training exposure for every model in Table~\ref{tab:result_tradition}: $\checkmark$ indicates use of the official training partition or a rewritten derivative, $\times$ indicates no reported use in the public training recipe, and ``?'' indicates undisclosed checkpoint-level exposure.}

\begin{table*}[p]
\centering
\scriptsize
\setlength{\tabcolsep}{2.5pt}
\caption{\revadd{Benchmark-training exposure for every model reported in Table~\ref{tab:result_tradition}.}}\label{tab:exposure_table2}
\begin{tabular}{@{}>{\revaddcolor}p{0.17\textwidth}*{4}{>{\revaddcolor\centering\arraybackslash}p{0.065\textwidth}}>{\revaddcolor}p{0.46\textwidth}@{}}
\toprule
\textbf{Model} & \textbf{VQA-RAD} & \textbf{SLAKE} & \textbf{Path\allowbreak VQA} & \textbf{PMC-VQA$^{*}$} & \textbf{Documented training stage / exposure basis}\\
\midrule
\multicolumn{6}{l}{\revadd{\textbf{Medical multimodal VLMs}}}\\
BiomedGPT & NAN & NAN & NAN & NAN & The paper reports task-specific fine-tuning on VQA-RAD, SLAKE, and PathVQA; the Table~\ref{tab:result_tradition} checkpoint is unspecified.\\
Med-R1-2B & $\times$ & $\times$ & $\times$ & $\times$ & GRPO training uses an open-access OmniMedVQA split; the four official training partitions are not listed.\\
MedVLM-R1-2B & NAN & NAN & NAN & NAN & GRPO uses a pooled set containing all four benchmarks; per-source split provenance is undisclosed.\\
HealthGPT-M3 & $\checkmark$ & $\checkmark$ & $\checkmark$ & $\times$ & Stage~3 visual instruction fine-tuning includes VQA-RAD, SLAKE, and PathVQA; PMC-VQA is not listed.\\
BioMedix2-8B & $\checkmark$ & $\checkmark$ & $\checkmark$ & $\times$ & Multimodal instruction alignment uses restructured VQA-RAD, SLAKE, and PathVQA training data.\\
MedGemma-4B-IT & $\checkmark$ & $\checkmark$ & $\times$ & $\times$ & Medical post-training data explicitly include VQA-RAD and SLAKE; PathVQA and PMC-VQA are not listed.\\
HuatuoGPT-V-7B & $\times$ & $\times$ & $\times$ & $\times$ & PubMedVision instruction tuning followed by zero-shot benchmark evaluation.\\
Lingshu-7B & $\checkmark$ & $\checkmark$ & $\checkmark$ & $\checkmark$ & Medical multimodal instruction tuning includes all four datasets.\\
Med-Flamingo & $\times$ & $\times$ & $\times$ & $\times$ & Continued medical pretraining; the evaluation recipe lists no target-split training.\\
RadFM & $\checkmark$ & $\checkmark$ & $\times$ & $\checkmark$ & Visual-instruction training includes VQA-RAD, SLAKE, and PMC-VQA; PathVQA is not listed.\\
LLaVA-Med-7B & $\times$ & $\times$ & $\times$ & $\times$ & The released PubMed-derived checkpoint is evaluated zero-shot.\\
+ PubMedVision & $\times$ & $\times$ & $\times$ & $\times$ & PubMedVision medical instruction tuning followed by zero-shot evaluation.\\
HuatuoGPT-Vision-34B & $\times$ & $\times$ & $\times$ & $\times$ & PubMedVision medical instruction tuning followed by zero-shot evaluation.\\
Hulu-Med-7B & $\checkmark$ & $\checkmark$ & $\checkmark$ & $\checkmark$ & Mixed-modality instruction tuning includes the four benchmark training sets or rewritten derivatives.\\
\midrule
\textbf{MultiViewDx-8B-AN} & $\checkmark$ & $\checkmark$ & $\checkmark$ & $\checkmark$ & Stage~3 human-annotated fine-tuning uses all four official training partitions.\\
\bottomrule
\end{tabular}
\par\noindent
\revadd{$^{*}$The PMC-VQA column records official QA training-partition exposure; Sec.~\ref{sec:evaluation_pipeline} describes the shared PMC/PMC-OA provenance.}
\end{table*}

\begin{table*}[p]
\centering
\scriptsize
\setlength{\tabcolsep}{2.5pt}
\begin{tabular}{@{}>{\revaddcolor}p{0.20\textwidth}*{4}{>{\revaddcolor\centering\arraybackslash}p{0.065\textwidth}}>{\revaddcolor}p{0.48\textwidth}@{}}
\toprule
\textbf{Model} & \textbf{VQA-RAD} & \textbf{SLAKE} & \textbf{Path\allowbreak VQA} & \textbf{PMC-VQA$^{*}$} & \textbf{Training stage in the Table~\ref{tab:finetune_trainset} setting}\\
\midrule
LLAVA-v1.5-LLAMA3-8B & $\checkmark$ & $\checkmark$ & $\checkmark$ & $\checkmark$ & Benchmark-specific supervised fine-tuning on each official training split.\\
LLAVA\_Med-8B & $\checkmark$ & $\checkmark$ & $\checkmark$ & $\checkmark$ & Benchmark-specific supervised fine-tuning on each official training split.\\
HuatuoGPT-Vision-8B & $\checkmark$ & $\checkmark$ & $\checkmark$ & $\checkmark$ & Benchmark-specific supervised fine-tuning on each official training split.\\
MultiViewDx & $\checkmark$ & $\checkmark$ & $\checkmark$ & $\checkmark$ & Benchmark-specific supervised adaptation on the corresponding official training split.\\
\bottomrule
\end{tabular}
\caption{\revadd{Benchmark-training exposure for every model in Table~\ref{tab:finetune_trainset}.}}
\label{tab:exposure_table3}
\end{table*}

\section{Additional Data Construction Examples}
\revmod{This section contains the JAMA case and paired ablation outputs, the view-removal result, benchmark-exposure tables, a human-annotated case in the MultiViewDx format, and a retrieval-guided synthesis example.}

\subsection{Human-Annotated Case Normalization}
\label{apx:human_case_example}

Table~\ref{case_eurorad} shows an expert-authored clinical case after being organized into the MultiViewDx format.
The example illustrates the information fields used in our human-annotated branch, including image caption, clinical history, image findings, discussion, and instruction-style supervision.
This supports the data construction section in the main paper by showing how expert-authored cases provide evidence-linked supervision rather than only image-answer pairs.

\subsection{Retriever-Guided Synthetic Construction}
\label{apx:retriever_example}

Table~\ref{retrieve_case} provides an example of the retrieval-guided synthesis process.
Without retrieved context, GPT-4o produces a generic image description and remains uncertain about the target structure.
With retrieved clinical cases, the generation is conditioned on related medical evidence and becomes more specific to the cardiac finding.
This example supports the data construction section by illustrating why retrieval is used in the GPT-4o synthesis branch: it provides contextual medical evidence for generating more grounded instruction data.

\clearpage
\onecolumn

\begingroup
\small
\setlength{\tabcolsep}{0pt}
\renewcommand{\arraystretch}{1.08}
\setlength{\arrayrulewidth}{0.7pt}

\begin{longtable}{|>{\raggedright\arraybackslash}p{0.97\textwidth}|}
\caption{JAMA case study illustrating a clinical prioritization error.}
\label{tab:case1} \\[2.0ex]
\hline

\noindent
\hspace*{2mm}
\smash{\raisebox{1.25ex}{\casebadge{Clinical Prioritization Error}}}

\begin{minipage}[t]{0.49\linewidth}
\vspace{0pt}
\textbf{Question.}
A woman in her mid-20s presented with subacute bilateral vision loss that was worse in the left eye. Her medical history was remarkable for type 1 diabetes diagnosed at 16 years of age and proliferative diabetic retinopathy in both eyes that had been treated with panretinal photocoagulation 7 years earlier. She had undergone pars plana vitrectomy with endolaser to treat a tractional retinal detachment in her right eye 2 years before this presentation. She also had a history of hypertension and chronic kidney disease, and she was
\end{minipage}
\hfill
\begin{minipage}[t]{0.47\linewidth}
\vspace{0pt}
\centering
\includegraphics[width=0.95\linewidth]{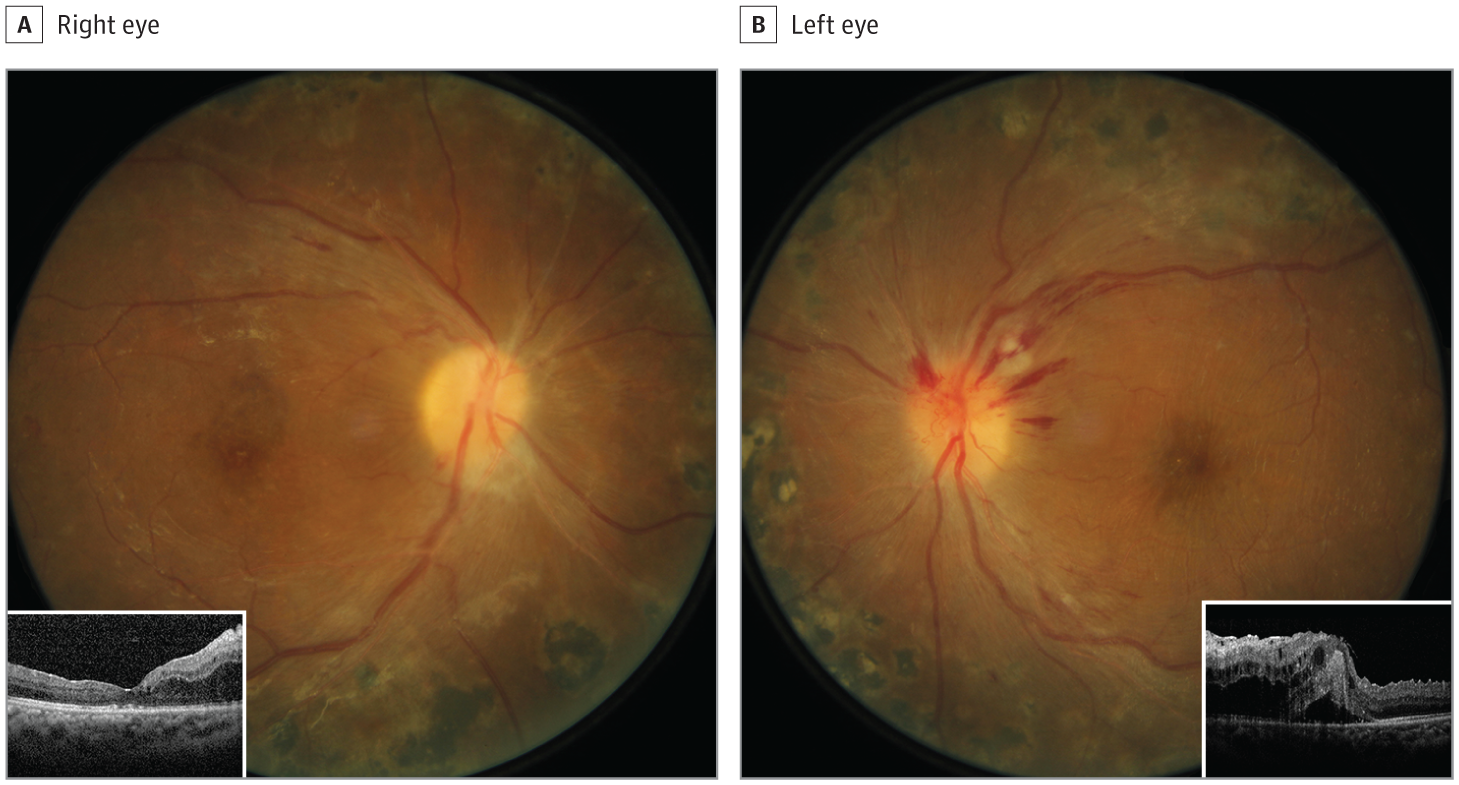}
\end{minipage}
\\[1mm]
 15 weeks into pregnancy. Visual acuity was 20/50 OD and 20/100 OS. Intraocular pressure was normal bilaterally, and no relative afferent pupillary defect was detected. Findings of an anterior segment examination were normal. The patient was in no apparent distress and denied any headache, chest pain, or focal weakness. Ophthalmoscopic examination (Figure) revealed mild optic nerve head edema that was greater in the left eye than the right eye with associated nerve fiber layer hemorrhage in the left eye. Nerve fiber layer infarctions, dot and blot hemorrhages, and lesions caused by panretinal photocoagulation also were seen bilaterally. Optical coherence tomography showed macular edema that involved the center of the macula in both eyes (Figure, inset). \\
\textbf{Choices:}\\
(A) Obtain a fluorescein angiogram;\\
(B) Determine blood glucose level and perform glycated hemoglobin test;\\
(C) Measure heart rate, respiratory rate, and blood pressure;\\
(D) Perform immediate computed tomography of the head.\\
\textbf{Correct Answer:}
\textcolor{green!60!black}{\textbf{(C)}} Measure heart rate, respiratory rate, and blood pressure.
\\
\textbf{Clinical Diagnosis:}
\textcolor{green!60!black}{\textbf{Malignant hypertension with papillopathy.}}
\\
\textbf{Diagnostic Reason:} \textcolor{red}{Malignant hypertension with papillopathy} \textcolor{red}{C. Measure heart rate, respiratory rate, and blood pressure} 
  The patient was found to have \textcolor{brown}{hypertension}, with a blood pressure of \textcolor{brown}{195/110 mm Hg}. Heart and respiratory rates were normal. Measurement of the arterial blood pressure may be performed rapidly in the clinic with a sphygmomanometer and is \textcolor{red}{essential to rule out malignant hypertension}, which is a \textcolor{red}{potentially life-threatening cause of vision loss}. 
  Although the differential diagnosis for \textcolor{red}{bilateral optic nerve edema} is broad, workup should always include assessment of blood pressure when appropriate, because a \textcolor{red}{hypertensive emergency} (also known as \textcolor{red}{malignant hypertension}) may cause substantial morbidity or mortality if not diagnosed and treated promptly. Findings may include \textcolor{brown}{macular star}, \textcolor{brown}{macular edema}, \textcolor{brown}{serous retinal detachment}, \textcolor{brown}{intraretinal hemorrhage}, and \textcolor{brown}{optic disc edema} with or without associated hemorrhage\textsuperscript{1,2}. Optic nerve head edema may occur  with systolic blood pressures as low as \textcolor{brown}{160 mm Hg}, with the median onset occurring at \textcolor{brown}{188 mm Hg}.\textsuperscript{3}\\ The macular edema associated with hypertensive retinopathy may be distributed more nasally, as was seen in this patient.\textsuperscript{4} 
  This patient had \textcolor{brown}{mild optic nerve edema} despite high systemic blood pressure and substantial macular edema. This less-pronounced optic nerve edema likely was attributable to \textcolor{brown}{optic nerve atrophy} at baseline. \textcolor{brown}{Proliferative diabetic retinopathy} and \textcolor{brown}{panretinal photocoagulation} can be associated with optic atrophy, and \textcolor{brown}{atrophic optic nerves} tend to become less edematous than healthy optic nerves.\textsuperscript{5-7}. Regarding the other choices above, a \textcolor{blue}{fluorescein angiogram (choice A)} would be expected to show leakage from the optic nerve and macula, but such findings are already available from the \textcolor{brown}{optical coherence tomography}, which showed \textcolor{brown}{intraretinal and subretinal fluid}. Although assessment of serologic levels of \textcolor{blue}{glucose and glycated hemoglobin (choice B)} is important in the management of diabetic retinopathy, results from such tests are not immediately available and have less bearing on the \textcolor{red}{immediate management} of the case (this patient’s glycated hemoglobin level was \textcolor{brown}{10.5\%} [to convert to a proportion of total hemoglobin, multiply by 0.01]). 
  Performing \textcolor{blue}{computed tomography (choice D)} would be a reasonable later step, especially if the patient’s blood pressure was found to be normal, to rule out an \textcolor{blue}{intracranial mass}. In addition to hypertension, other causes of optic nerve head edema exist. \textcolor{blue}{Papilledema} (when optic nerve head edema is secondary to \textcolor{blue}{increased intracranial pressure}) is often bilateral and may be caused by \textcolor{blue}{intracranial mass lesions}, \textcolor{blue}{meningitis}, \textcolor{blue}{cerebral venous thrombosis}, or \textcolor{blue}{intracranial hypertension}, or it may be idiopathic. Bilateral optic nerve head edema also may be caused by \textcolor{blue}{diabetic papillopathy} (which is a rare cause of optic disc edema, but which is bilateral in up to 50\% of cases\textsuperscript{8,9}) and \textcolor{blue}{toxic optic neuropathies}. \textcolor{blue}{Optic disc drusen} are frequently bilateral and may cause so-called \textcolor{blue}{pseudopapilledema}.\textsuperscript{10} 
  \textcolor{brown}{Severe hypertension resulting in acute retinopathy and optic nerve head edema} is considered a \textcolor{red}{hypertensive emergency} because of its association with \textcolor{red}{end-organ damage}, namely, damage to the \textcolor{brown}{retina and optic nerve}. \\
  This case serves as a reminder that, although the differential diagnosis for optic nerve head edema is broad, \textcolor{red}{immediately assessing the patient’s blood pressure} to evaluate for \textcolor{red}{hypertensive emergency} in the eye clinic may be \textcolor{red}{critical}, because this treatable condition could lead to \textcolor{red}{substantial morbidity or mortality} if diagnosis is delayed. 
  The patient was admitted to the medicine service for \textcolor{brown}{management of hypertension}. At an ophthalmology follow-up examination 3 months later, her blood pressure was \textcolor{brown}{128/73 mm Hg} and her visual acuity was \textcolor{brown}{20/40 OD} and \textcolor{brown}{20/60 OS}, close to her visual acuity before this acute event. On examination, the \textcolor{brown}{optic nerve head edema had resolved}, and \textcolor{brown}{optical coherence tomography showed resolution of macular edema}.\\
\hline

\rowcolor{gray!20}
\textbf{Prediction Rationale} \\
\hline
\textbf{MultiViewDx.}
\textcolor{green!60!black}{\textbf{C. Measure heart rate, respiratory rate, and blood pressure.}} 
\textcolor{red}{Malignant hypertension with papillopathy} \textcolor{red}{C. Measure heart rate, respiratory rate, and blood pressure}
  The key to the correct diagnosis is the presence of \textcolor{red}{bilateral optic neuropathy in a pregnant woman}.
  The patient’s history of \textcolor{blue}{hypertension and chronic kidney disease}, along with the findings of \textcolor{blue}{bilateral optic neuropathies}, suggested a \textcolor{blue}{systemic etiology}.
  A fluorescein angiogram (choice A) would not be the next step because the patient was \textcolor{blue}{asymptomatic} and had \textcolor{blue}{no signs of retinal ischemia}.
  A glycated hemoglobin test (choice B) would not be indicated because the patient had \textcolor{blue}{no signs of diabetic retinopathy}.
  A computed tomography scan of the head (choice D) would not be appropriate because the patient had a \textcolor{blue}{normal examination} and \textcolor{blue}{no signs of intracranial hemorrhage}. 
  Besides hypertension, other causes of optic nerve head edema exist. Papilledema, which is optic nerve head edema secondary to increased intracranial pressure, is often bilateral and may result from intracranial mass lesions, meningitis, cerebral venous thrombosis, \textcolor{red}{intracranial hypertension, or it may be idiopathic.}
  Bilateral optic nerve head edema \\can also be caused by diabetic papillopathy—a rare cause of optic disc edema that is bilateral in up to 50\% of cases¹—and toxic optic neuropathies. Optic disc drusen, which are frequently bilateral, may lead to so-called pseudopapilledema.
  \textcolor{brown}{Severe hypertension causing acute retinopathy and optic nerve head edema} is considered a \textcolor{brown}{hypertensive emergency} due to its association with \textcolor{brown}{end-organ damage}, specifically affecting the retina and optic nerve.
  This case emphasizes that, although the differential diagnosis for optic nerve head edema is extensive, \textcolor{brown}{promptly assessing the patient's blood pressure} in the eye clinic to evaluate for a hypertensive emergency is \textcolor{brown}{critical}.
  \textcolor{brown}{Immediate evaluation is crucial} because this treatable condition could lead to \textcolor{brown}{significant morbidity or mortality} if diagnosis is delayed.
  The patient was admitted to the medical service for \textcolor{brown}{hypertension management}. At an ophthalmology follow-up examination three months later, her blood\\
    pressure was 128/73 mm Hg, and her visual acuity improved to 20/40 in the right eye (OD) and 20/60 in the left eye (OS), nearing her visual acuity before this acute event. Upon examination, the optic nerve head edema had resolved, and optical coherence tomography showed a resolution of the macular edema.
\\
\hline

\rowcolor{gray!20}
\textbf{Baseline Error Pattern}
\\
\hline
\textbf{Claude3-Opus.}
This case presents several concerning features that make \textcolor{red}{measuring vital signs the most urgent next step}:
  \textcolor{red}{Bilateral optic nerve edema (worse in left eye)} in a \textcolor{red}{young pregnant woman} with:
  \begin{itemize}
    \item \textcolor{red}{Type 1 diabetes}
    \item \textcolor{red}{Hypertension}
    \item \textcolor{red}{Chronic kidney disease}
    \item \textcolor{red}{Active pregnancy (15 weeks)}
  \end{itemize}

The key concern here is \textcolor{red}{POSTERIOR REVERSIBLE ENCEPHALOPATHY SYNDROME (PRES)}, which is:
  \begin{itemize}
    \item \textcolor{blue}{A neurological condition characterized by vasogenic edema}
    \item \textcolor{brown}{Commonly associated with severe hypertension}
    \item \textcolor{brown}{More common in pregnant women}
    \item \textcolor{blue}{Can present with visual symptoms and optic nerve edema}
    \item \textcolor{brown}{Can be life-threatening if not treated promptly}
  \end{itemize}
  
  Why other options are less appropriate:
  \begin{itemize}
    \item \textcolor{blue}{A (Fluorescein angiogram): Contraindicated in pregnancy}
    \item \textcolor{blue}{B (Blood glucose/HbA1c): While important for diabetic management, not the most urgent concern given the presentation}
    \item \textcolor{blue}{D (Head CT): While eventually needed if PRES is suspected, vital signs should be checked first as they'll guide immediate management}
  \end{itemize}
  \\
  The vital signs are crucial because:
  \begin{itemize}
    \item They will help determine if this is a hypertensive emergency requiring immediate intervention.
    \item They're non-invasive and provide immediate critical information
  \end{itemize}
  
  The findings will guide the urgency and direction of further workup
  \begin{itemize}
    \item If severely elevated blood pressure is found, immediate treatment could prevent serious complications for both mother and fetus
    \item This represents a potential medical emergency where checking vital signs is the most appropriate first step to guide immediate management decisions.
  \end{itemize}
  \\\hline
\textbf{HuatuoGPT-Vision-34B.}
The patient, who has a history of \textcolor{red}{type 1 diabetes} and is currently \textcolor{red}{15 weeks pregnant}, presents with \textcolor{red}{optic nerve head edema, macular edema, and hemorrhages in both eyes}; therefore, it is crucial to \textcolor{blue}{assess her blood glucose levels and perform a glycated hemoglobin (HbA1c) test to evaluate and manage her glycemic control status}, as this evaluation is essential to \textcolor{brown}{identify any acute changes or complications related to her pregnancy and existing diabetes}, since \textcolor{brown}{poor glycemic control can lead to rapid progression of diabetic retinopathy, exacerbate ocular conditions, and pose significant risks to both maternal and fetal health—including the potential for congenital anomalies, preeclampsia, and worsening of chronic kidney disease}—thus, timely\\ \textcolor{blue}{ assessment and adjustment of her diabetes management plan are vital to prevent further vision loss and ensure the well-being of both the mother and the developing fetus}.
\\
\hline
\textbf{HuatuoGPT-Vision-7B.}
\textcolor{red}{B. Determine blood glucose level and perform glycated hemoglobin test}
  Reasoning: The patient has a history of \textcolor{red}{type 1 diabetes}, which is known to cause \textcolor{brown}{diabetic retinopathy leading to vision loss} as seen in the images provided (\textcolor{brown}{optic nerve head edema with hemorrhage}). The presence of \textcolor{brown}{macular edema} also suggests \textcolor{brown}{ongoing or poorly controlled diabetes} affecting her eyes.
  \textcolor{blue}{A fluorescein angiogram would be useful for assessing vascular abnormalities but may not directly address the underlying issue related to poor glycemic control}.
  Measuring heart rate, respiratory rate, and blood pressure are important general assessments but do not specifically relate to the ophthalmic findings presented here.
  Immediate computed tomography of the head might be considered if there was suspicion of other complications like intracranial pathology; however, it's less relevant given the current clinical context focused on eye examination results linked to diabetes management issues.
\\
\hline

\textbf{GPT-o-mini.} Correct Answer: \textcolor{red}{C. Measure heart rate, respiratory rate, and blood pressure}. 
  Reason: \textcolor{red}{Measuring blood pressure is critical} in this scenario to identify \textcolor{red}{malignant hypertension}, a potentially \textcolor{brown}{life-threatening condition} that can lead to \textcolor{brown}{vision loss through optic nerve and retinal damage}. \textcolor{blue}{Immediate assessment of blood pressure helps rule out hypertensive emergencies}, allowing for \textcolor{brown}{timely intervention}.\\

\hline
\rowcolor{gray!20}

\rowcolor{gray!20}
\textbf{Expert Analysis}
\\
\hline

\begin{tabularx}{\linewidth}{@{}>{\bfseries\raggedright\arraybackslash}p{3.2cm}X@{}}
\textbf{Model} & \textbf{Answer and Main Behavior} \\
\midrule

MultiViewDx &
\textcolor{green!60!black}{\textbf{Correct: (C).}}
Correctly links bilateral optic nerve edema, pregnancy, hypertension, and chronic kidney disease to the need for immediate vital-sign assessment and blood pressure measurement.
\\[0.5ex]

Claude3-Opus &
\textcolor{green!60!black}{\textbf{Correct: (C).}}
Correctly prioritizes vital signs, but frames the concern more broadly as a hypertensive neurologic emergency.
\\[0.5ex]

GPT-4o-mini &
\textcolor{green!60!black}{\textbf{Correct: (C).}}
Selects the correct action but provides a short rationale with limited supporting evidence.
\\[0.5ex]

HuatuoGPT-Vision-34B &
\textcolor{red!70!black}{\textbf{Incorrect: (B).}}
Focuses on diabetes and glycemic control, missing the more urgent need to rule out hypertensive emergency.
\\[0.5ex]

HuatuoGPT-Vision-7B &
\textcolor{red!70!black}{\textbf{Incorrect: (B).}}
Attributes the eye findings mainly to diabetic retinopathy and fails to prioritize blood pressure assessment.
\end{tabularx}
\\[-0.5mm]
\hline
MultiViewDx and Claude3-Opus accurately recognized the urgency of measuring vital signs to assess for malignant hypertension in this patient, who presents with bilateral optic nerve edema and a history of hypertension and chronic kidney disease during pregnancy. Their reasoning reflects a proper understanding of the need for immediate intervention to prevent serious complications. On the other hand, Huatuo-GPT-Vision-34B and Huatuo-GPT-Vision-7B focused on assessing glycemic control by selecting to determine blood glucose levels and perform a glycated hemoglobin test. While managing diabetes is important, they failed to prioritize the immediate life-threatening condition suggested by the patient's symptoms, thus overlooking the critical need to rule out a hypertensive emergency. GPT-4o-mini could get the correct answer and some key points but lack lots of detail evidence to prove it.
\\
\hline

\end{longtable}
\endgroup

\begingroup
\small
\setlength{\tabcolsep}{0pt}
\renewcommand{\arraystretch}{1.08}
\setlength{\arrayrulewidth}{0.7pt}

\begin{longtable}{|>{\raggedright\arraybackslash}p{0.97\textwidth}|}
\caption{Human-annotated clinical case normalized into the MultiViewDx evidence-linked schema.}\label{case_eurorad} \\[1.5ex]
\hline

\noindent
\hspace*{2mm}
\smash{\raisebox{1.2ex}{\casebadge{Human Annotated Sample Case}}}

\vspace{1.5ex}

\makebox[\linewidth][c]{\includegraphics[width=0.95\linewidth]{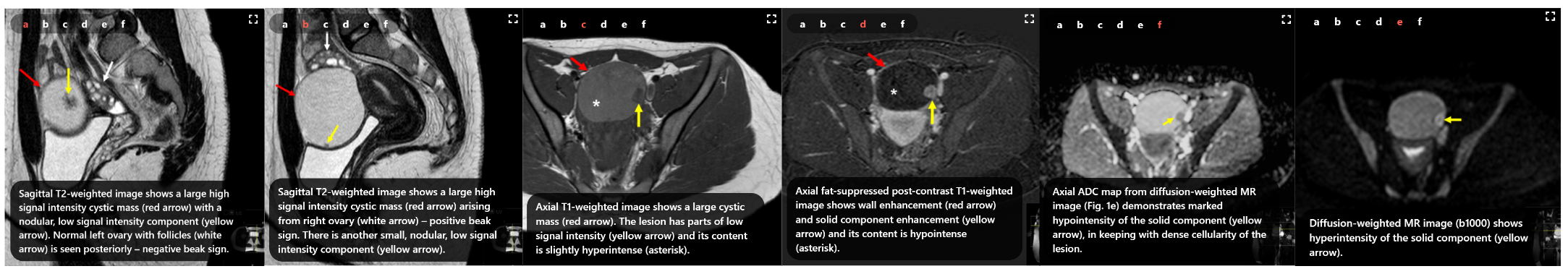}}
\\[1mm]
\hline

\textbf{Image Caption:} 1. Sagittal T2-weighted image shows a large high signal intensity cystic mass (red arrow) with a nodular, low signal intensity component (yellow arrow). Normal left ovary with follicles (white arrow) is seen posteriorly – negative beak sign. 2. Sagittal T2-weighted image shows a large high signal intensity cystic mass (red arrow) arising from right ovary (white arrow) – positive beak sign. There is another small, nodular, low signal intensity component (yellow arrow). 3. Axial T1-weighted image shows a large cystic mass (red arrow). The lesion has parts of low signal intensity (yellow arrow) and its content is slightly hyperintense (asterisk). 4. Axial fat-suppressed post-contrast T1-weighted image shows wall enhancement (red arrow) and solid component enhancement (yellow arrow) and its content is hypointense (asterisk). 5. Diffusion-weighted MR image (b1000) shows hyperintensity of the solid component (yellow arrow). 6. Axial ADC map from diffusion-weighted MR image (Fig. 1e) demonstrates marked hypointensity of the solid component (yellow arrow), in keeping with dense cellularity of the lesion.
\\[1mm]
\hline

\textbf{Clinical History:} A 21-year-old G0P0 woman with no medical history was referred to our institution for a sonographically detected cystic right adnexal mass. She has a history of pelvic discomfort without other complaints. Physical examination was normal. Laboratory findings were also normal except for an elevated\\ CA 125 65.2 U/mL (normal <35.0).
\\[1mm]
\hline

\textbf{Image Findings:} MRI examination revealed a cystic tumour arising from the right ovary with 7.5 cm. On T2-weighted images, the signal intensity of the cyst content was high and two small nodular peripheral solid components were detected, adhering to its internal wall, with low signal (Fig. 1a, b). The normal left ovary was present with follicles (Fig. 1a).
 On pre-contrast T1-weighted images, the mass exhibited slightly high signal intensity (Fig. 1c). On contrast-enhanced fat-suppressed T1-weighted images, wall enhancement and solid component enhancement were detected (Fig. 1d). Finally, the ADC map (Fig. 1f) from diffusion-weighted image (Fig. 1e) demonstrates marked hypointensity of the solid component, in keeping with its dense cellularity. Surgical excision was proposed and accepted by the patient. The histopathological investigation revealed a typical ovarian serous borderline tumour.
\\[1mm]
\hline

\textbf{Discussion:} Borderline ovarian tumours are uncommon ovarian neoplasms, intermediate between benign and malignant types, corresponding to 5\% of all epithelial ovarian tumours. [1, 2] Serous borderline tumour represents the most common type of borderline tumours arising in the ovary, and typically, it is confined to the adnexa and presents an indolent course. [3] However, up to 6.8\% of these tumours can progress to low grade serous carcinoma. [3] Serous borderline tumours are divided into typical (90\%) and borderline tumours with micro-papillary patterns (5\%–10\%). [4] These neoplasms usually present as bilateral adnexal masses with more\\ proliferation of papillary projections than do benign cystadenomas, they are often seen in younger patients, and laboratory findings show the serum CA-125 level mildly elevated. [2, 3, 5, 6] The peak age of presentation is 45 years. [5] Small tumours usually do not cause symptoms and are often detected as an incidental finding on sonography. [7] Larger or more advanced neoplasms might cause pain or pelvic discomfort. The diagnosis of this type of tumour is based on histopathological examination. As they are staged using the same ovarian cancer staging of malignant ovarian neoplasms [5], MRI plays a crucial rule in this evaluation. There are no pathognomonic imaging features of borderline ovarian tumours and they frequently resemble a benign lesion. Nevertheless, some pathologic and MR imaging studies suggested that large papillary projections are highly suggestive of borderline or malignant neoplasms. [1, 8] The preferred treatment is bilateral oophorectomy. Fertility sparing procedures such as unilateral oophorectomy or cystectomy are related with a higher incidence of recurrence when compared with bilateral oophorectomy (10-20\% vs. approximately 5\% for radical surgery). [3, 4] Young age (<30 years) has also been found to be a risk factor for recurrent disease in patients treated conservatively. As in this case, patients should be instructed to complete child-bearing within a short time frame following initial diagnosis, due to the high risk of recurrence. [3] The most important feature in predicting an adverse outcome in serous borderline tumours is the presence of serous lesions involving the peritoneum (so-called invasive implants). [3, 4] Tumours without invasive implants, as in this case, have been considered to be nonaggressive tumours, associated with a good prognosis and high survival rate. Follow-up using ultrasound imaging is mandatory, with particular attention paid to the remaining ovary in conservatively treated young patients. [4] Written informed patient consent for publication has been obtained.
\\
\hline

\textbf{MultiViewDx(only a sample for one image):}
\\[0.5mm]

\textbf{User:}What type of imaging technique is shown in Fig. 1a of the provided image?
\\
\textbf{GPT:}The image shows a T2-weighted MRI scan.
\\[0.5mm]

\textbf{User:}Where is the cystic tumour located according to the image?
\\
\textbf{GPT:}The cystic tumour is arising from the right ovary
\\[0.5mm]

\textbf{User:}What is the size of the cystic tumour?
\\
\textbf{GPT:}The cystic tumour is 7.5 cm in size.
\\[0.5mm]

\textbf{User:}What does the red arrow in the image indicate?
\\
\textbf{GPT:}The red arrow indicates the normal left ovary with follicles.
\\[0.5mm]

\textbf{User:}What does the yellow arrow in the image point to?
\\
\textbf{User:}The yellow arrow points to one of the small nodular peripheral solid components adhering to the internal wall of the cystic tumour.
\\[0.5mm]

\textbf{User:}What characteristic of the cyst content is shown by the high signal intensity in the T2-weighted image?
\\
\textbf{User:}The high signal intensity indicates that the cyst content is fluid-filled.
\\[0.5mm]

\textbf{User:}What is shown by the asterisk (*) in the image?
\\
\textbf{User:}The asterisk (*) indicates the cystic tumour arising from the right ovary.
\\
\hline

\end{longtable}
\endgroup

\begingroup
\small
\setlength{\tabcolsep}{0pt}
\renewcommand{\arraystretch}{1.08}
\setlength{\arrayrulewidth}{0.7pt}

\begin{longtable}{|>{\raggedright\arraybackslash}p{0.97\textwidth}|}
\caption{Retriever-guided synthesis example comparing GPT-4o output with and without retrieved clinical context.}\label{retrieve_case} \\[1.5ex]
\hline

\noindent
\hspace*{2mm}
\smash{\raisebox{1.2ex}{\casebadge{Retriever-Guided Example}}}

\vspace{1.5ex}
\\

\begin{minipage}[t]{0.34\linewidth}
\vspace{0pt}
\centering
\includegraphics[width=4cm]{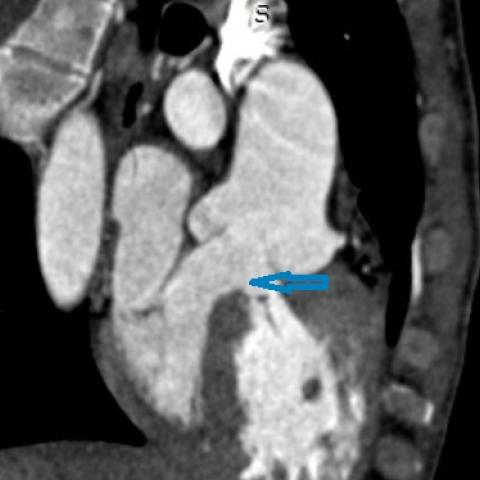}
\end{minipage}
\hfill
\begin{minipage}[t]{0.63\linewidth}
\vspace{0pt}

\vspace{1mm}
\textbf{Prompt:} Generate the Image description based on the image. \\
\textbf{GPT-4o:} The image is a sagittal CT scan, likely of the thoracic or abdominal region. The scan shows various soft tissue structures, including major organs. The blue arrow points toward a specific structure, possibly a blood vessel or part of an organ. This could indicate an abnormality such as a mass, an aortic aneurysm, or an area of vessel narrowing or dissection. In the background, major anatomical landmarks like the heart, aorta, and vertebral column are visible, as well as some portions of the liver or spleen. The areas of high contrast indicate dense structures, such as bones, while softer tissues are seen in varying shades of gray, highlighting different densities. This scan could be used to evaluate conditions like vascular abnormalities, tumors, or other pathologies in the chest or upper abdominal region. Further clinical information would be necessary to confirm the exact diagnosis.
\end{minipage}
\\[1mm]
\hline
\rowcolor{gray!20}
\textbf{Prompt(with retriever)} 
\\
\hline

\begin{minipage}[t]{0.50\linewidth}
\vspace{0pt}
\centering
\includegraphics[width=0.48\linewidth]{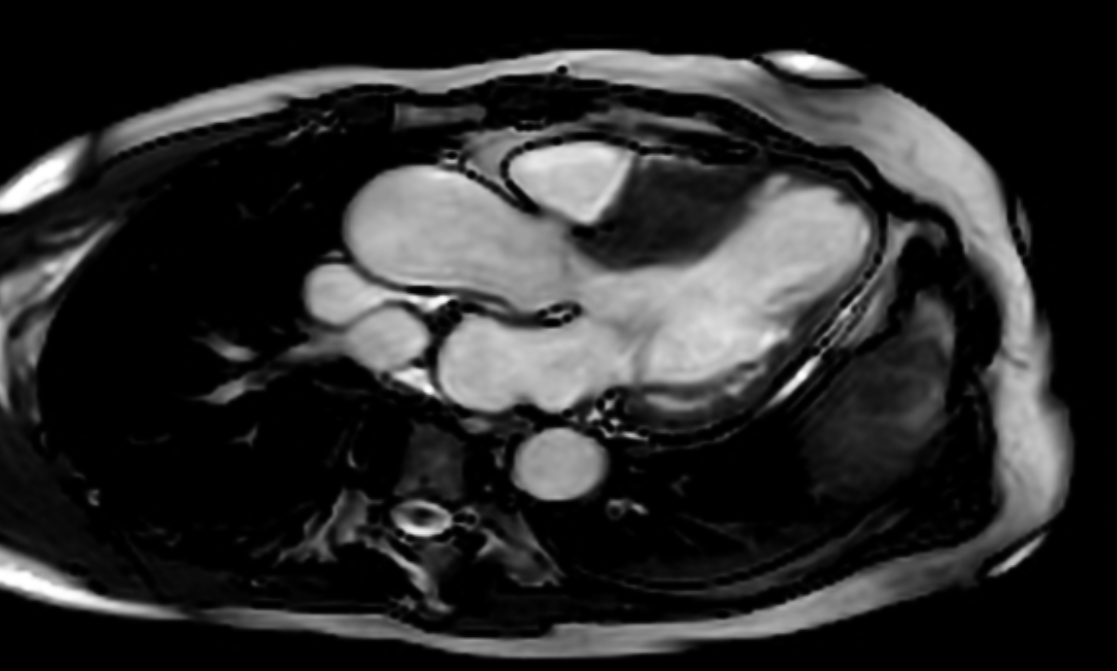}
\hfill
\includegraphics[width=0.48\linewidth]{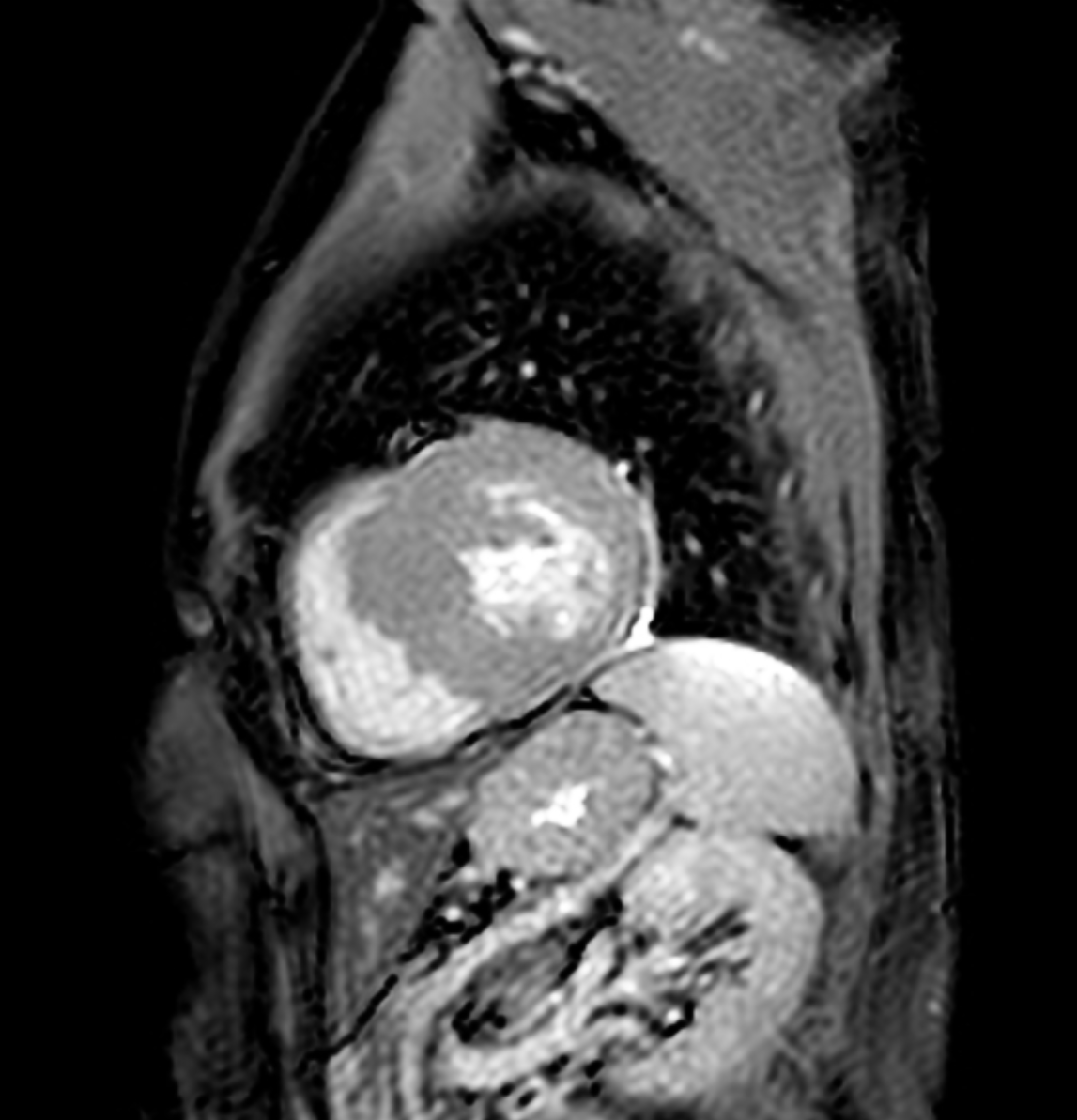}
\end{minipage}
\hfill
\begin{minipage}[t]{0.47\linewidth}
\vspace{0pt}

\vspace{1mm}

\textbf{Prompt:} Generate the Image description based on the image. Here are some medical guidelines and Clinical cases for you. \\
\textbf{Case Discussion:} A case of asymmetric hypertrophic cardiomyopathy with mild subaortic stenosis and partial systolic anterior motion of the mitral valve. Since the pressure gradient obtained by phase contrast imaging was rather low, echocardiographic reevaluation and/or follow-up was recommended. The case also demonstrates a typical pattern of myocardial 
\end{minipage}
\\[1mm]

fibrosis in the basal and midventricular septal segments at the anterior and posterior right ventricular insertion points.\\

\vspace{1mm}

\textbf{GPT-4o:} The image is a coronal oblique CT scan of the thoracic region, focused on the heart and its surrounding structures. The blue arrow indicates a subaortic ventricular septal defect (VSD), measuring approximately 2.7 cm. This congenital heart defect refers to an abnormal opening in the interventricular septum (the wall separating the left and right ventricles of the heart), located just below the aortic valve.
  The surrounding anatomy includes parts of the heart chambers, likely showing the left and right ventricles, as well as nearby vascular structures such as the aorta.
  The defect appears to allow abnormal communication between the ventricles, which could lead to altered blood flow dynamics, potentially causing symptoms such as shortness of breath, fatigue, or heart failure if left untreated.\\
\hline
\end{longtable}
\endgroup

\end{document}